\documentclass[final,5p,times,twocolumn,sort&compress]{elsarticle}
\usepackage{amsmath,amsfonts}
\usepackage{algorithmic}
\usepackage{algorithm}
\usepackage{array}
\usepackage[caption=false,font=normalsize,labelfont=sf,textfont=sf]{subfig}
\usepackage{textcomp}
\usepackage{stfloats}
\usepackage{url}
\usepackage{verbatim}
\usepackage{graphicx}
\usepackage{svg}
\usepackage{times}
\usepackage[utf8]{inputenc}
\usepackage{url}
\usepackage{amssymb}
\usepackage{amstext}
\usepackage{amsmath}
\usepackage{booktabs}
\usepackage{tabularx}
\usepackage{threeparttable}
\usepackage{multirow, makecell} 
\usepackage{rotating}
\usepackage{nicefrac} 
\usepackage{setspace}
\usepackage[english]{babel}
\usepackage{soul}

\usepackage{xurl}
\usepackage{hyperref}
\usepackage{cleveref}
\usepackage{xcolor}
\usepackage{hyperref}

\hypersetup{colorlinks=true, linkcolor=blue, anchorcolor=red, citecolor=blue, filecolor =red, pdfauthor=author}

\newcommand{\bb}{\color{black}}
\usepackage{utfsym}
\newcommand{\cmark}{\usym{1F5F8}}
\newcommand{\xmark}{\scalebox{0.95}{\usym{2613}}}
\usepackage{color, colortbl}

\usepackage{pgfplots}
\pgfplotsset{compat=1.18}
\usepackage{layouts}
\usepackage{comment}
\usepackage{threeparttable}
\usepackage{bm}
\usepackage{enumitem}
\usepackage{setspace}
\usepackage{chngcntr}
\counterwithin{table}{section} 

\usepackage{times}
\usepackage{balance}  
\usepackage{soul}               
\usepackage{url}
\usepackage{textcomp}
\usepackage{mathrsfs}
\usepackage{dcolumn}
\usepackage{color}
\usepackage{array}
\usepackage{textcomp}
\usepackage{siunitx}

\definecolor{Gray}{gray}{0.9}

\usepackage{longtable}
\usepackage{pdflscape}
\usepackage{lineno}

\definecolor{pv}{HTML}{4F6272}       % blu per PV
\definecolor{weath}{HTML}{C25E5E}     % rosso per Weather
\definecolor{fused}{HTML}{E3B04B}  % arancione per Fused
\definecolor{forecast}{HTML}{5C8D4C}% verde per Weather Forecast

\DeclareSIUnit{\day}{\text{\ensuremath{\mathrm{day}}}}
\usepackage{acronym}
\acrodef{res}[RES]{Renewable Energy Source}
\acrodef{pv}[PV]{photovoltaic}
\acrodef{ai}[AI]{Artificial Intelligence}
\acrodef{nwp}[NWP]{Numerical Weather Prediction}
\acrodef{dl}[DL]{Deep Learning}
\acrodef{bilstm}[BiLSTM]{bi-directional short-term memory}
\acrodef{lstm}[LSTM]{Long-Short-Term Memory}
\acrodef{gru}[GRU]{Gated Recurrent Unit}
\acrodef{bigru}[BiGRU]{bi-directional GRU}
\acrodef{dni}[DNI]{Direct Normal Irradiance}
\acrodef{dhi}[DHI]{Diffuse Horizontal Irradiance}
\acrodef{ghi}[GHI]{Global Horizontal Irradiance}

\addto\extrasenglish{%
}

\addto\extrasenglish{%
}

\newcommand{\refappendix}[1]{\hyperref[#1]{\ref*{#1}}}

\journal{}

\begin{document}

\begin{frontmatter}


%% use the ead command for the email address,ets

\title{MATNet: Multi-Level Fusion Transformer-Based Model for Day-Ahead PV Generation Forecasting}


\author[aff1]{Matteo Tortora\corref{cor1}} %% Author name
%\ead{m.tortora@unicampus.it}

\author[aff2]{Francesco Conte}
\author[aff1]{Gianluca Natrella}
\author[aff3]{Paolo Soda}

\address[aff1]{Department of Naval, Electrical, Electronics and 
Telecommunications Engineering,\\ University of Genoa, Via all’Opera Pia 11a, 16145 Genoa, Italy}

\address[aff2]{Unit of Innovation, Entrepreneurship \& Sustainability, \\ Department of Engineering, University Campus Bio-Medico of Rome\\ Via Alvaro del Portillo 21, 00128 Rome, Italy}

\address[aff3]{Unit of Artificial Intelligence and Computer Systems \\ Department of Engineering, University Campus Bio-Medico of Rome\\  Via Alvaro del Portillo 21, 00128 Rome, Italy}

\cortext[cor1]{Corresponding author: Matteo Tortora, E-mail: matteo.tortora@unige.it, Address: Via all’Opera Pia 11a, 16145 Genoa, Italy.}

%% Abstract
\begin{abstract} 
Accurate forecasting of renewable generation is crucial to facilitate the integration of \acp{res} into the power system. Focusing on \ac{pv} units, forecasting methods can be divided into two main categories: physics-based and data-based strategies, with \ac{ai}-based models providing state-of-the-art performance. 
However, while these \ac{ai}-based models can capture complex patterns and relationships in the data, they ignore the underlying physical prior knowledge of the phenomenon. 
Therefore, in this paper, we propose MATNet, a novel transformer-based multimodal architecture for multi-step day-ahead \ac{pv} power generation forecasting.
The model is fed with historical \ac{pv} data and historical and forecast weather data through a multi-level joint fusion approach, employing a soft-attention mechanism at multiple fusion stages.
We evaluate the effectiveness of MATNet on the Ausgrid benchmark dataset, where it significantly outperforms various baseline models, achieving an RMSE of 0.0445, corresponding to a relative improvement of approximately 65\% compared to the best-performing baseline method.
The analysis is further enriched by a comprehensive set of ablation studies, a sensitivity analysis on missing data, which highlights MATNet’s resilience to input degradation, a cross-site zero-shot generalization evaluation on five external PV datasets, demonstrating MATNet’s robustness under significant domain shifts, and an assessment of the model’s computational complexity, confirming its favorable balance between predictive accuracy and computational efficiency. 
These results highlight MATNet’s potential as a reliable and efficient solution to facilitate the integration of \ac{pv} energy into the power grid.
The code is available at~\url{https://github.com/arco-group/MATNet}.
\end{abstract}

%% Keywords

\end{frontmatter}



\section{Introduction}
\label{sec:intro}
The integration of renewable energy into power systems is steadily increasing 
due to the combination of climate change and the consequent need to reduce greenhouse gas emissions.
The shift towards \acp{res} is crucial to a sustainable, affordable, accessible, clean, and low-carbon future, reducing polluting emissions and dependence on fossil fuels.  
In this regard, \acp{pv} energy certainly is one of the most mature \ac{res} technologies, called to play a crucial role in accomplishing various climate protection goals~\cite{InternationalSolar, International2022Solar, zhou2022artificial}.  

Compared to fossil fuel-derived energy, green energy is substantially more sustainable. However, their inherent intermittent nature does not guarantee constant production flows, causing imbalances in the power system that ultimately limit large-scale adoption~\cite{meenal2022weather,simeunovic2021spatio}. \ac{pv} energy generation forecasting could help resolve these imbalances and uncertainties, facilitating the general integration of \ac{res} into the power systems~\cite{simeunovic2021spatio,wan2015photovoltaic,gupta2021pv,aslam2021survey,kaur2022energy}.
%Accurate forecasting of \ac{pv} production is therefore crucial for realizing the full potential of \ac{pv} systems.
%Indeed, \ac{pv} forecasting emerges as an essential stakeholder providing grid operators and energy traders with valuable insights and decision-making information for optimising maintenance strategies, planning the development of new plants, mitigating operational and management challenges, and improving economic returns on investment.

Accurately forecasting \ac{pv} production is essential to realizing the full potential of \ac{pv} systems and providing grid operators and energy traders with valuable insights and decision-making information to optimize maintenance strategies, plan the development of new plants, mitigate operational and management challenges, and improve economic returns on investment~\cite{alkhayat2021review,conteCER,bianchi,wan2015photovoltaic}.
For this reason, several methods for forecasting \ac{pv} energy production have recently been developed~\cite{iheanetu2022solar}. Currently, \ac{pv} forecasting methods can be divided into physics-based and data-based strategies~\cite{gupta2021pv}. 
The former, also known as \ac{nwp}, are mathematical models that simulate complex systems to predict how the atmosphere will evolve. These models are based on fluid dynamics and thermodynamic principles and use a combination of observations from weather stations, satellites, radars, and other sources to predict atmospheric dynamics. 
However, the main drawback of these methods is their lack of flexibility, as they require intensive knowledge about the considered phenomena and can be expensive in terms of time and resources. 
Additionally, they require extensive computer resources to run effectively~\cite{wan2015photovoltaic,chai2020robust}.
The latter (i.e., data-based approaches) can be further split into statistics- and Artificial Intelligence (AI)-based methods.
On the one hand, statistical methods include well-known approaches like linear regression, Auto-Regressive Moving Average or Auto-Regressive Integrated Moving Average, which
aim to establish the mathematical framework governing the data-generation process~\cite{barker2020machine}.
However, despite their prevalent application in the literature and while some incorporate nonlinear approaches, they may not be optimal for uncovering complex nonlinear relationships~\cite{januschowski2020criteria}. 
On the other hand, \ac{ai}-based models have dominated research in recent years due to their ability to discover complex nonlinear relations, deal with unstructured data, and superior performance.
Compared with the former methods, they allow for easier modeling without requiring prior knowledge of the phenomenon's dynamics~\cite{li2023tcn}.
In particular, \ac{dl} has shown promising results in forecasting PV energy production due to its ability to generalize and automatically extract abstract representations, and recurrent neural networks (RNNs) as well as convolutional neural networks (CNNs) are the most used \ac{dl}-based architectures in \ac{pv} power forecasting~\cite{aslam2021survey}.
However, despite their advantages, \ac{dl} methods ignore the domain-specific physical knowledge that could enhance their interpretability and accuracy. 
For instance, in \ac{pv} systems, critical factors such as weather conditions, which directly influence energy production, are not explicitly modeled. 
\ac{nwp}-based methods, in contrast, excel at incorporating this knowledge, albeit with limitations in scalability and flexibility.
This dichotomy highlights the need for frameworks that can integrate diverse data modalities, effectively leveraging the complementary strengths of these approaches. 

Multimodal learning offers a promising solution to address these challenges by integrating heterogeneous data sources to enhance predictive performance beyond the limitations of stand-alone modalities, demonstrating notable success across diverse fields, ranging from medical diagnostics to natural language processing~\cite{tortora2023radiopathomics, furia2023exploring,ayllon2025,omar2025}, despite the absence of formal theoretical proof.
Multimodal learning can be implemented at different levels through three fusion strategies: early fusion, joint fusion, and late fusion~\cite{baltruvsaitis2018multimodal,ramachandram2017deep}.
Early fusion involves combining raw data from multiple modalities at the input stage, allowing the model to learn joint feature representations from the beginning. While straightforward, this approach faces challenges when handling inputs with varying scales or temporal resolutions, requiring the model to account for these differences during training. 
Joint fusion, on the other hand, integrates information at an intermediate level, combining features extracted separately from each modality. This approach preserves the unique characteristics of each modality while learning shared representations, making it particularly effective for capturing complex, nonlinear interdependencies between data sources. 
In contrast, late fusion aggregates outputs from models trained independently on each modality at the decision-making stage. While this strategy is simpler and can reduce overfitting risks, it may miss opportunities to capture deeper interdependencies between modalities during training.
For an in-depth discussion of these fusion techniques, see~\cite{baltruvsaitis2018multimodal}.

The adoption of multimodal learning can facilitate the development of robust and accurate PV forecasting models by integrating the complementary strengths of physics- and data-driven approaches.
Nevertheless, its application in PV forecasting remains relatively underexplored~\cite{li2023tcn, bai2022deep}.
\bb
To fill this gap, in this work, we propose MATNet, a novel multimodal self-attention-based architecture for multi-step, day-ahead \ac{pv} power production forecasting, combining the advantages of a \ac{dl} approach with the a priori knowledge of the phenomenon provided by physics-based models.
The contributions of this work are summarized as follows:
\begin{itemize}
    \item We propose MATNet, a novel transformer-based multimodal architecture for multi-step day-ahead \ac{pv} generation forecasting, incorporating three data modalities: PV production data, weather history conditions, and weather forecast. 
    \item We introduce a multi-level joint fusion framework integrating heterogeneous data sources through a multi-stage soft-attention mechanism. 
    By leveraging the temporal distinction between historical and forecast data, the framework dynamically adjusts contributions at different stages, effectively capturing cross-modal dependencies and enhancing predictive accuracy.
    \item Inspired by interpolation techniques from language modeling, we propose a dense interpolation module
    that generates concise representations while preserving temporal structure from high-dimensional attention-based outputs. This layer adaptively learns weights for interpolating hidden states, capturing the temporal dynamics of time-series data and enhancing the flexibility of the final representation.
    \item We evaluate the model's effectiveness on the Ausgrid benchmark dataset~\cite{ausgrid}, comparing its performance against thirteen baseline models, including standard statistical approaches, machine learning methods, and deep neural networks.
    The analysis is further extended through ablation studies, a sensitivity analysis on missing data, a cross-site zero-shot generalization evaluation on five external PV datasets, and a comprehensive assessment of the model’s computational complexity.
    
\end{itemize}

The rest of this manuscript is organized as follows: ~\autoref{sec:relatedwork} presents a review of related work. ~\autoref{sec:materials} introduces the dataset, overviewing the pre-processing adopted. ~\autoref{sec:methods} describes the details of our approach, whilst \autoref{sec:result} describes the experimental setup and discusses the results. Finally, ~\autoref{sec:conclusion} provides concluding remarks.

\section{Related Work}
\label{sec:relatedwork}
This section will focus only on work related to the Ausgrid dataset~\cite{ausgrid} which contains the historical individual energy production data of 300 residential \ac{pv} solar units. In~\autoref{subsec:ausgrid}, we will analyze this dataset in further detail.

In~\cite{fentis2020machine} the authors proposed an ensemble approach for 24-hour-ahead \ac{pv} generation forecasting.
The idea behind this is to break down the \ac{pv} production time series into sub-time sequences, so that each of these sub-time series collects \ac{pv} power records for the same time and all days of the year, creating a time series for every 30 minutes of production.
Therefore, their approach consists of an ensemble of simple univariate models based on Least Squares Support Vector Regression (LsSVR), each of which predicts the \ac{pv} output of each 30 minutes of the following day.
%The authors validated their approach using a publicly available \ac{pv} power database provided by Ausgrid, an Australian electric utility.
They demonstrated that the proposed approach is generic and any machine learning algorithm for regression can be used.
Despite the model giving good results for \emph{clear} and \emph{partly cloudy} days, it fails to handle the \emph{cloudy/overcast} type of day.

Building on the limitations of traditional machine learning models, in~\cite{kaur2021bayesian} the authors introduce a Bayesian probabilistic method incorporating Bidirectional Long Short-Term Memory (BiLSTM) networks. 
This model addresses uncertainties in \ac{pv} generation data and model parameters, achieving multi-step ahead forecasts with improved probabilistic accuracy. 
Results demonstrated its effectiveness over other methods by evaluating probabilistic metrics like Pinball loss and Winkler score, underscoring the model's robustness.
Building on this, in a subsequent work~\cite{kaur2022bayesian} the authors proposed an improved Bayesian \ac{bilstm}-based \ac{dl} technique integrating an alpha-beta divergence, specifically designed to handle outliers in \ac{pv} generation data. 
The results demonstrate that this improved Bayesian \ac{bilstm} with alpha-beta divergence outperforms standard Bayesian \ac{bilstm} and other benchmark methods for multi-step ahead forecasting in terms of lower error values, leading to better error reduction and robustness against data anomalies.
Lastly, the same authors proposed a further enhancement in~\cite{kaur2023vae}, which combines Bayesian BiLSTM with variational autoencoders to tackle the high dimensionality in weight parameters. 
This approach for \ac{pv} forecasting reduces computational overhead by compressing the dimensionality of the parameter space. 
The model's efficiency was demonstrated with significant reductions in computational time and improved forecasting accuracy, thus providing a scalable solution for renewable energy forecasting with uncertain, high-dimensional data.

These approaches, while effective in certain conditions, reveal notable limitations under adverse weather conditions tied to their underlying unimodal methodologies. 
Unimodal models restrict their focus to a single data source, such as historical PV power production, without incorporating complementary data sources. 
To fill these gaps, we propose a novel multimodal architecture for multi-step, day-ahead PV forecasting, integrating heterogeneous sources such as historical PV production data, weather history, and weather forecasts within a unified framework that leverages advanced attention mechanisms to capture intricate temporal and cross-modal dependencies among these data streams.
This multimodal approach not only enhances the model’s ability to adapt to diverse weather conditions but also improves accuracy and robustness in forecasting under varying weather scenarios.
    \begin{figure}
    \centering
    \includegraphics[width=\columnwidth]{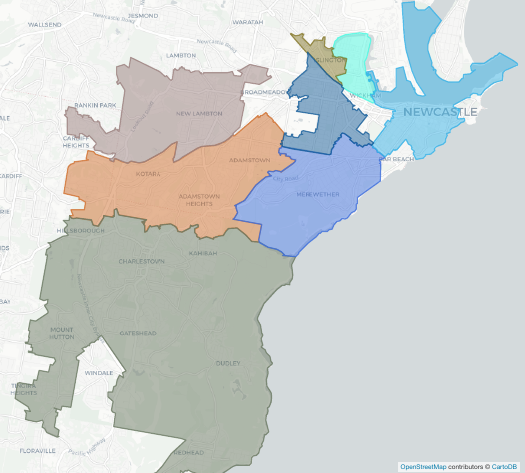}
    \caption{Map of the territorial areas containing the 26 \ac{pv} units considered in this study, generated using the PyTrack library~\cite{tortora2022pytrack}. Each colored area corresponds to a distinct geographical region associated with one of the eight zip codes.
    The study area falls within the Cfa (humid subtropical) climate zone according to the Köppen–Geiger classification.}
    \label{fig:maps}
\end{figure}

\section{Materials}
\label{sec:materials}
In this work, we utilize three primary data sources for multi-step \ac{pv} power production forecasting: Ausgrid, OpenWeatherMap, and Solcast.
Ausgrid is the electricity distributor for New South Wales, Australia, and provides a dataset on historical electricity demand and \ac{pv} power generation. 
OpenWeatherMap is an online weather data service that provides real-time, forecast, and historical weather data.
Lastly, Solcast is an online solar data service providing forecasts, live and historical solar irradiance, as well as weather data, which is freely available for public research purposes or implementation in household systems.
In the following, we will describe in further detail the characteristics and features of these datasets, as well as the pre-processing steps used in this study.

\begin{table*}
\centering
\caption{Weather attributes extracted from OpenWeather API and Solcast.}
\label{tab_feat_abs}
%\resizebox{\columnwidth}{!}{%
\begin{tabular}{l c l}%{l c p{0.2\textwidth}}
\toprule
\textbf{Attribute} & \textbf{UoM} & \textbf{Description} \\ 
\midrule
Temperature & \si{\kelvin} & Temperature \\ 
Pressure & \si{\hecto\pascal}	& Atmospheric pressure (on the sea level) \\ 
Humidity &  \(\%\) & Humidity \\ 
Dew point & \si{\kelvin} & Temperature at which condensation occurs \\
Wind speed & \si{\meter\per\second} & Wind speed 
\\
Wind deg & deg & Wind direction \\ 
Clouds all & \si{\percent} & Cloudiness \\ 
Rain 1h	& \si{\milli\meter} & Rain volume for the last hour \\ 
Weather description & - & Categorical weather conditions \\ 
\ac{dni} & \si{\watt\per\meter\squared} & Direct Normal Irradiance 
\\
\ac{dhi} & \si{\watt\per\meter\squared} & Diffuse Horizontal Irradiance
\\
\ac{ghi} & \si{\watt\per\meter\squared} & Global Horizontal Irradiance
\\\bottomrule
\end{tabular}
%}
\end{table*}

\subsection{Ausgrid}
\label{subsec:ausgrid}
We utilized the ``Solar home electricity dataset" provided by Ausgrid~\cite{ausgrid} as the primary source of data.
This electricity distributor provider owns, maintains, and operates the electricity distribution network in Sydney, the Central Coast, and the Hunter regions of New South Wales, Australia.
The dataset contains energy generation (in kWh) for 300 residential rooftop solar \ac{pv} units recorded directly from the \ac{pv} inverter over the period from 1 July 2010 to 30 June 2013 with a sampling time of 30 minutes. 
To ensure data integrity, we selected only 26 out of 300 households with \ac{pv} units spread over a 75-square-kilometer area and located in 8 different zip codes within the Newcastle region (as illustrated in~\autoref{fig:maps}), which falls within the Cfa (humid subtropical) climate zone according to the Köppen–Geiger classification; a comprehensive list of the included customers is reported in~\refappendix{app:ausgridcustomers}.
While this filtering ensures high-quality and consistent measurements, it may also introduce a mild selection bias toward PV systems with more stable operation, slightly limiting the dataset’s representativeness under more heterogeneous conditions.

\begin{figure*}
    \centering
    \includegraphics[width=0.8\linewidth]{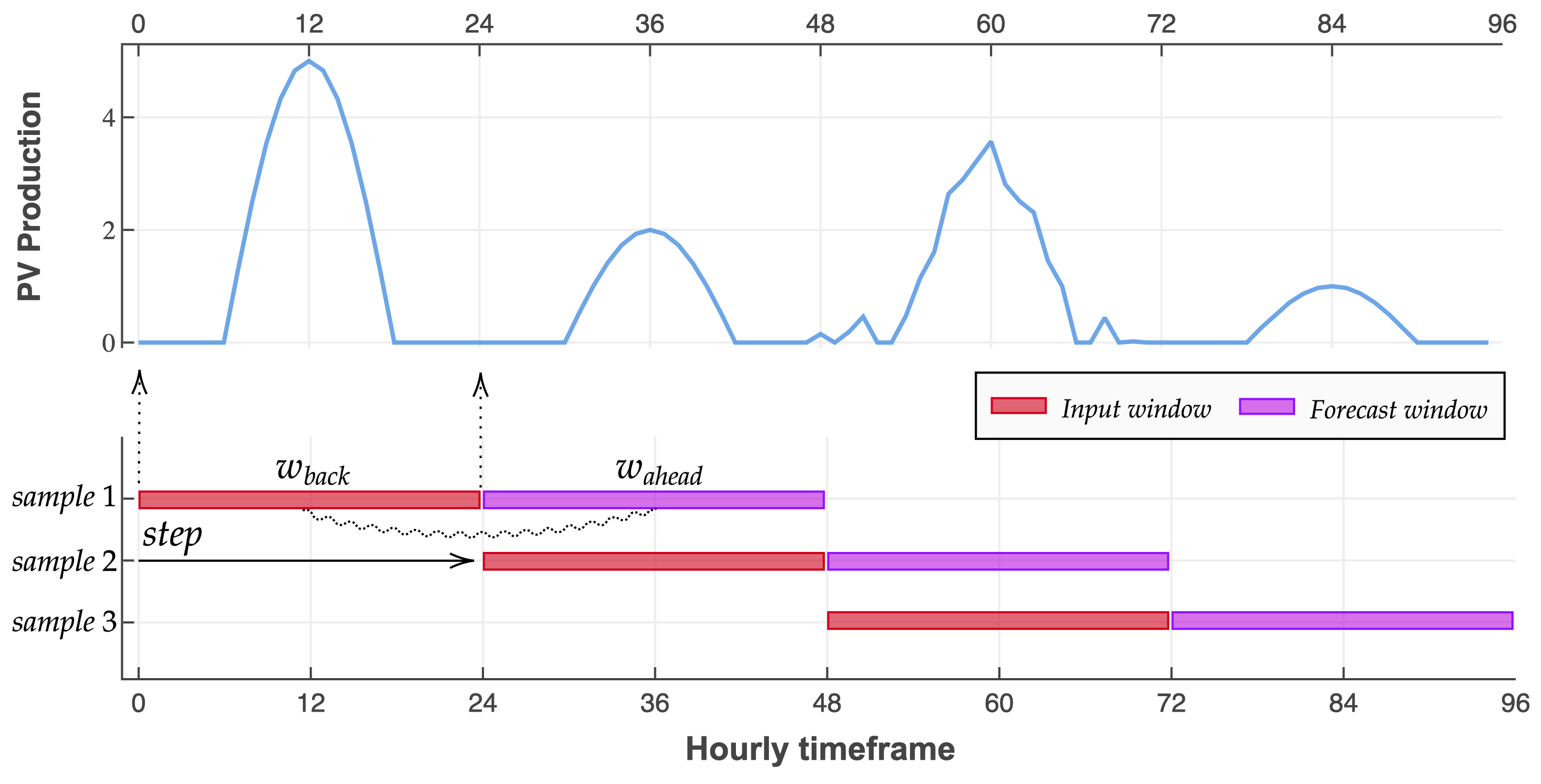}
    \caption{
    Sliding window scheme for day-ahead PV forecasting. Red and purple bars denote the input (\(w_{back}\)) and forecast (\(w_{ahead}\)) windows, respectively, shifted over time with a fixed \(step\).}
    \label{fig:sliding_window}
\end{figure*}

We pre-processed the data as follows.
Let \( X = \{ x_i\left(t_m\right) \mid i = 1, \dots, N; \ m = 1, \dots, T \} \) denote the collection of \ac{pv} energy time series for \( N \) households over \( T \) time steps, 
%and \( t_1, \dots, t_T \) correspond to the original time steps 
with a sampling interval \( \Delta t = 30 \) minutes.
Here \(x_i\left(t_m\right)\) represents the energy generated over the half-hour interval \(\left(t_{m-1},t_m\right]\).
First, we aggregated the series to an hourly sampling frequency by summing the two half-hour values that fall within each hour, a choice motivated by the need to align with practical energy forecasting applications such as day-ahead scheduling and market bidding~\cite{perera2022multi,entsoe_2022_report,epexspot}.\footnote{A sensitivity analysis across different input granularities is reported in~\refappendix{app:granular}. Results show that MATNet maintains stable performance across resolutions, confirming the suitability of the 1-hour granularity.}
Let \(\tau_k=t_{2k-2}\) denote the start time of the \(k\)-th hour and let \(\tilde{T}=\lfloor T/2\rfloor\) be the number of hourly steps.
The aggregated hourly series for the \(i\)-th household is:
\begin{equation}
\label{eq:coarser}
    \tilde{x}_i\left(\tau_k\right)=\sum_{j=1}^2=x_i\left(t_{2k-2+j}\right), \quad k=1,2,\dots,\tilde{T}.
\end{equation}
Each value \(\tilde{x}_i\left(\tau_k\right)\) therefore corresponds to the energy generated in the hour \(\left(\tau_k, \tau_k + 1 h\right]\).
Next, we aggregated the subsampled \ac{pv} facilities' data across all households by summing the energy production of each household at every timestamp:
\begin{equation*}
\tilde{P}_{agg}\left(\tau_k\right) = \sum_{i=1}^{N} \tilde{x}_i\left(\tau_k\right), \quad k = 1, 2, \dots, \tilde{T}
\end{equation*}
To improve numerical stability during model training, we normalized the aggregated time series by the total installed capacity of all households, represented as the sum of the kilowatt-peak (\( kWp \)) of each household, yielding the hourly specific yield: 

\begin{equation*}
\overline{P}\left(\tau_k\right) = \frac{\tilde{P}_{agg}\left(\tau_k\right)}{\sum_{i=1}^{N} kWp_i}, \quad k = 1, 2, \dots, \tilde{T}
\end{equation*}
This normalization not only ensures that all values are scaled between 0 and 1, but also leads to better conditioning of the problem and faster learning during the training phase, facilitating a more efficient optimization process~\cite{lecun2002efficient}.
Finally, to prepare the data for model training, we extracted samples using a sliding window parameterized by two factors: the width of the sliding block (\( w \)) and the step size (\(step\)) between windows. 
Each window serves as an input sample,
paired with target values corresponding to the aggregated \ac{pv} energy production for the subsequent 24-hour period.
A schematic illustration of the sliding window extraction process is provided in~\autoref{fig:sliding_window}.
The specific values of the parameters are detailed in~\autoref{subsec:training}.
For the sake of simplicity in the subsequent text, we denote the pre-processed PV power production time series as 
$\mathbf{pv} \in \mathbb{R}^{w_{\text{back}} \times 1}$, representing the historical PV power production over a look-back window of $w_{\text{back}}$ time steps.

\subsection{Weather Conditions}
We utilized as the primary source for meteorological data OpenWeatherMap service~\cite{openweather}, which provides weather data for any location using an \ac{nwp} proprietary model. 
It is fed with 82,000 weather stations spread globally, radars, and weather satellites.

In addition to the meteorological data obtained from OpenWeatherMap, we incorporated solar radiation data from Solcast~\cite{2019Global}. By incorporating the following measurements we got an in-depth insight into the availability and intensity of solar energy at a specified location:
\begin{itemize}[noitemsep]
\item \ac{dni}: It represents the direct irradiance received on a surface held perpendicular to the sun, providing information about solar radiation intensity when the sunlight reaches the surface directly without any obstructions.
\item \ac{dhi}: It refers to the diffuse irradiance received on a horizontal surface. It quantifies the scattered radiation caused by atmospheric factors such as clouds, haze, and pollution.
\item \ac{ghi}: This measurement represents the irradiance received on a horizontal surface. It takes into account both direct sunlight and diffuse sky radiation and is crucial for assessing the solar energy potential of a location.
\end{itemize}

\autoref{tab_feat_abs} summarizes the climate attributes provided by both the OpenWeatherMap API and Solcast considered in this work.
Unlike all other variables which are numerical features, the \emph{weather description} attribute is a categorical feature with \(22\) distinct levels, listed in depth in~\refappendix{app:categoricalwxfeatures}.
In order to convert these categorical values into numeric ones we apply a one-hot encoding, as no ordinal relation exists for these variables.

As for the power production data, we generated samples through a sliding window process.
Since OpenWeatherMap and Solcast do not provide historical weather forecast data for the considered period, in this work, we simulated the required forecast data by employing historical weather data as a proxy for the subsequent 24-hour time steps.
It is worth noting that this substitution does not introduce major inconsistencies because weather data contains two types of uncertainty that effectively emulate forecast scenarios. 
The first is the modeling error in the underlying NWP models, which reflects the unavoidable inaccuracies in simulating complex atmospheric dynamics. 
The second is the spatial discrepancy, arising because PV units are distributed across an area of approximately 75 square kilometers. 
This distribution introduces variability as the weather data reflects conditions averaged over a broader spatial region rather than specific localized measurements.
These uncertainties align with the imperfections typically observed in forecast weather data, thereby validating the use of historical data for simulation purposes.
To further enhance the robustness and realism of the simulation, and in particular to mimic the random variations often present in weather forecasts, we introduce a 5\% Gaussian noise to the historical weather data in all experiments conducted in this study. 
Furthermore, we perform a sensitivity analysis across different noise levels in the range \([0, 20\%]\), with a step of 5\%. 
This analysis, detailed in~\refappendix{app:sens_noise}, shows that MATNet maintains stable performance across all noise levels.
While this simulation provides a reasonable approximation of real forecast conditions, we acknowledge that it cannot fully replicate the behavior of actual operational forecasts.
%At the end of this pre-processing process for the meteorological data, we obtained a multimodal time series consisting of 33 attributes.
At the end of this pre-processing process for the meteorological data, we obtained two multimodal time series consisting of 33 attributes:  
$\mathbf{hw} \in \mathbb{R}^{w_{\text{back}} \times 33}$, representing the historical weather data spanning a look-back window of $w_{\text{back}}$ time steps, 
and $\mathbf{fw} \in \mathbb{R}^{w_{\text{ahead}} \times 33}$, corresponding to the forecasted weather data over a look-ahead window of $w_{\text{ahead}}$ time steps.

\begin{figure*}[!t]
    \centering
    \includegraphics[scale=0.65]{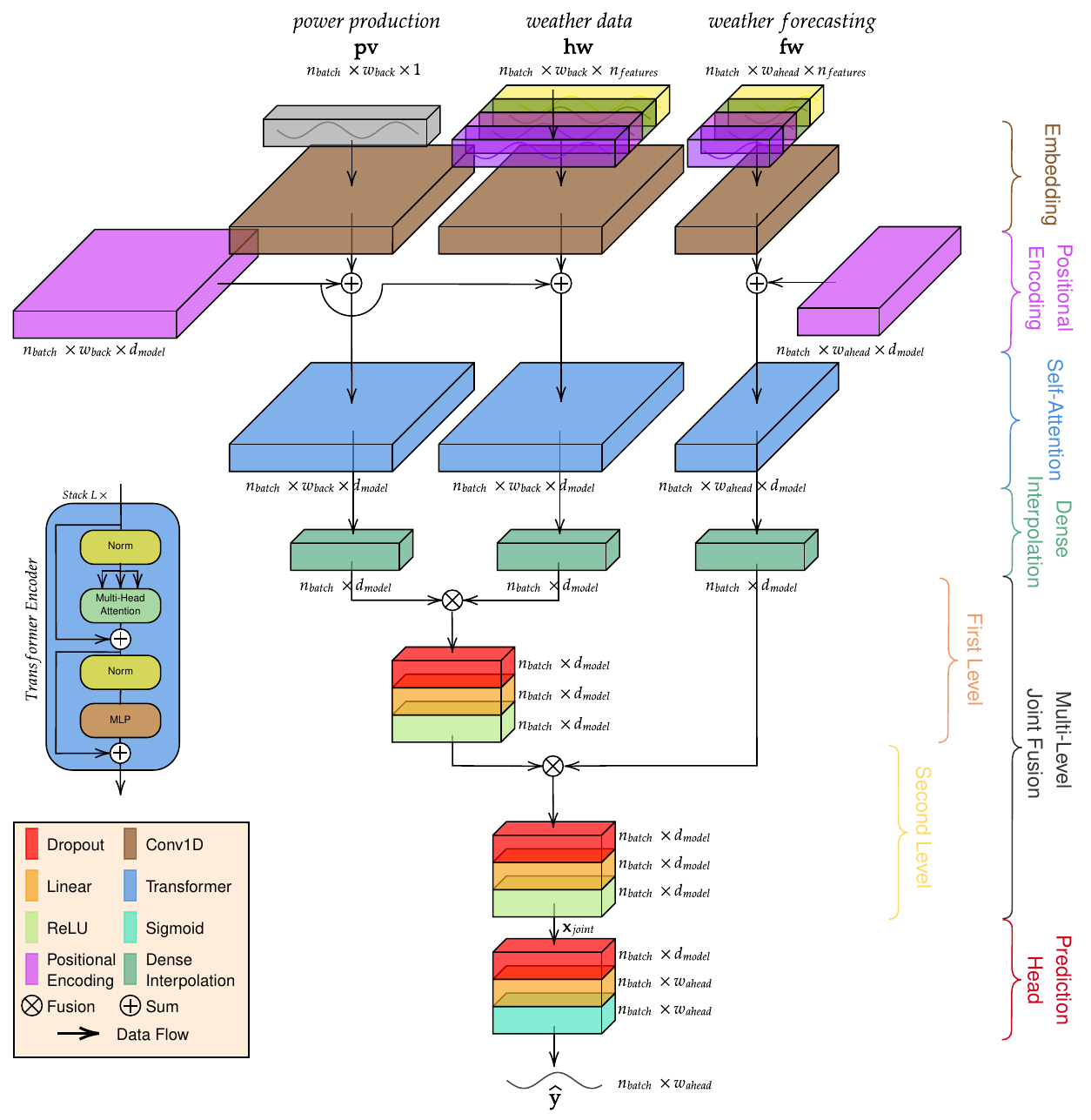}
    \caption{Schematic architecture of our proposed MATNet.}
    \label{fig:architecture}
\end{figure*}

\section{Proposed Architecture}
\label{sec:methods}
This section presents MATNet, our multi-level fusion and self-attention transformer-based model for multimodal multi-step day-ahead \ac{pv} forecasting, fed with historic \ac{pv} production and historical and forecasted weather data. 
The proposed architecture, depicted in~\autoref{fig:architecture}, is composed of several key components: embedding modules that project input time series into high-dimensional representations, positional encoding layers to encode temporal order, self-attention modules to highlight relevant features within the sequence, dense interpolation layers that consolidate temporal structures, and a multi-level joint fusion framework that dynamically combines representations from different modalities using a soft-attention mechanism, culminating in the final multimodal vector for \ac{pv} power prediction.
The following sections provide an in-depth description of these components, progressing systematically from the top to the bottom of the architecture, explaining their roles and interactions in achieving the model's objectives.

\subsection{Embedding Module}
Our architecture accepts as input a \(3\)-tuple of multimodal time series of the type \(\langle \mathbf{pv}, \mathbf{hw}, \mathbf{fw} \rangle\), where \(\mathbf{pv} \in \mathbb{R}^{w_{back} \times 1}\), \(\mathbf{hw} \in \mathbb{R}^{w_{back} \times 33}\) and \(\mathbf{fw} \in \mathbb{R}^{w_{ahead} \times 33}\) represents the historical \ac{pv} power production, the historical weather data and the forecast weather data of the next 24-hour time steps to be predicted, respectively, with \(w_{back}\) and \(w_{ahead}\) denoting the look-back and look-ahead windows.
The specific values of \(w_{back}\) and \(w_{ahead}\) are detailed in~\autoref{subsec:training}.
The first step in our architecture is an embedding module that maps the multivariate time series into a high-dimensional representation space to facilitate the sequence modeling.
This projection enables the model to capture complex relationships within the data that may not be evident in the original input. 
To this end, we implement a 1D convolutional layer as an embedding module to project the input data to a \(d_{model}\)-dimensional space:
\begin{equation*}
    \mathbf{o}_i = \mathbf{b}_i + \sum_{k=0} ^ {C_{in}-1} {\mathbf{w}_{i, k} \ast \mathbf{in}_{i, k}} \quad \forall i \in \left\{ 0,\dots,C_{out}-1 \right\}
\end{equation*}
where \(\mathbf{w}\in \mathbb{R}^{1\times size}\) is the convolution kernel, \(C_{in}\) and \(C_{out}\) are the input and the output channels, \(\mathbf{b}\) is the bias term, \(\ast\) is the sliding dot product operator (brown block in~\autoref{fig:architecture}).

When working with \ac{pv} production data (\(C_{in}=1\)), we use a convolutional layer with a kernel size of 3 to capture short-term temporal dependencies while preserving the local structure of the data.
However, when analyzing weather data (\(C_{in}=33\)), we use a kernel size of 1, allowing the convolutional layer to focus on capturing dependencies across different time series attributes without considering temporal information.
In both cases, the number of output channels is set to \(C_{out}=d_{model}\).
The specific value of \(d_{model}\) is detailed in~\autoref{subsec:training}.

In conclusion, the output of the embedding module produces multimodal time series with the same number of time steps as the input, but in a higher-dimensional representation with \(d_{model}\) attributes.
These %higher-dimensional 
representations are then used as input for the attention mechanism described in~\autoref{subsec:attn}.

\subsection{Positional Encoding}
As our architecture does not contain any recurrence, we include a positional encoding layer to incorporate the relative or absolute position of the time steps within the input-embedded sequence. 
This is particularly important for sequential data, as the ordering of the elements can carry valuable information allowing the model to better capture the temporal dependencies in the data and improve the performance of the network.

There are various choices for positional encodings, such as randomized lookup tables or learnable representations~\cite{gehring2017convolutional}.
Here we use sine and cosine functions of different frequencies as in~\cite{vaswani2017attention}, because they provide a smooth, continuous representation of positions and naturally encode relative distances between time steps, which is beneficial for capturing temporal relationships:
\begin{equation*}
\overrightarrow{p_t}^{(i)}:=\begin{cases}
\sin{\omega_k}\cdot t,  & \text{if \( i = 2k \) } \\
\cos{\omega_k}\cdot t, & \text{if \( i = 2k+1 \) }
\end{cases},
\end{equation*}
where:
\begin{equation*}
    \omega_k=\frac{1} {10000^{\frac{2k}{d_{model}}}}
\end{equation*}
and  \(\overrightarrow{p_t}^{(i)}\) represents the positional encoding for  the \(t\)-th time step in the input sequence along the \(i\)-th dimension, and \(d_{model}\) is the input embedding dimension.
The \(d_{model}\)-dimensional positional embedding is then added to the input embedding since they share the same dimension (light purple block in~\autoref{fig:architecture}).

\subsection{Self-Attention}
\label{subsec:attn}
Our MATNet architecture relies on self-attention mechanisms to process sequential input data. 
Self-attention, also known as intra-attention, allows the model to attend to different parts of the whole input sequence, weighing the importance of each component at each position. 
Unlike recurrent layers that process input data sequentially using recurrent connections and maintaining hidden states to propagate information through time, the self-attention mechanism processes the entire input sequence in parallel, weighing the importance of each element independently~\cite{vaswani2017attention, hochreiter1997long}.

An attention function maps a query and a set of key-value pairs to an output, and it is mathematically defined as ~\cite{vaswani2017attention}:
\begin{equation*}
    \text{Attention}\left(\mathbf{Q}, \mathbf{K}, \mathbf{V}\right) = \text{softmax}\left(\frac{\mathbf{Q}\mathbf{K}^T}{\sqrt{d_{model}}}\right)\mathbf{V}
\end{equation*}
where \(d_{model}\) is the dimension of the key vectors, and \(\mathbf{Q}\), \(\mathbf{K}\) and \(\mathbf{V}\) are all linear transformations of the input embedding with the added positional encodings.
We extended the attention mechanism using multi-head attention: instead of performing a single attention function with keys, values, and queries \(d_{model}\)-dimensional matrices, the attention mechanism is split into \(h\) distinct heads, each operating on an independent projection of the input data.
Specifically, the input embeddings are linearly projected into \(h\) lower-dimensional subspaces.
These \(h\)-projected representations fed the attention mechanisms in parallel, yielding \(d_v\)-dimensional output values allowing the model to attend to information from different representation sub-spaces at different positions (with \(d_v=d_{model}/h\)).
Each head can be interpreted to encode complementary details on different types of temporal dependencies~\cite{song2018attend}.
Finally, heads are concatenated and linearly projected to obtain the final representation:
\begin{equation*}
    \begin{split}
        \text{MultiHead}\left(\mathbf{Q}, \mathbf{K}, \mathbf{V}\right) & = \text{Concat}\left(head_1, \dots,head_h\right)\mathbf{W}^O \\
        \text{where } head_i & =\text{Attention}\left(\mathbf{Q}\mathbf{W}_i^Q, \mathbf{K}\mathbf{W}_i^K, \mathbf{V}\mathbf{W}_i^V\right)
\end{split}
\end{equation*}
where \(\mathbf{W}_i^Q \in \mathbb{R}^{d_{model} \times d_k}\), \(\mathbf{W}_i^K \in \mathbb{R}^{d_{model} \times d_k}\), \(\mathbf{W}_i^V \in \mathbb{R}^{d_{model} \times d_v}\), and \(\mathbf{W}^O \in \mathbb{R}^{hd_{v} \times d_{model}}\) are the learnable projections matrices for \(i\)-th head and for the output, respectively.
In this work, we use the same dimension for the queries, keys and values \(d_k=d_v=d_{model}/h\).
The specific value of \(h\) is detailed in~\autoref{subsec:training}.

The second block of the attention module is a simple feed-forward network downstream of the multi-head attention layer.
Besides, a normalization layer is applied before every block, whilst residual connections are after every block~\cite{dosovitskiy2020image, baevski2018adaptive}.

This attention module is stacked \(L\) times, and the final representations of the entire sequence are obtained from the last one (light blue block in~\autoref{fig:architecture}). 

\begin{figure}
    \centering
    \includegraphics[scale=0.85]{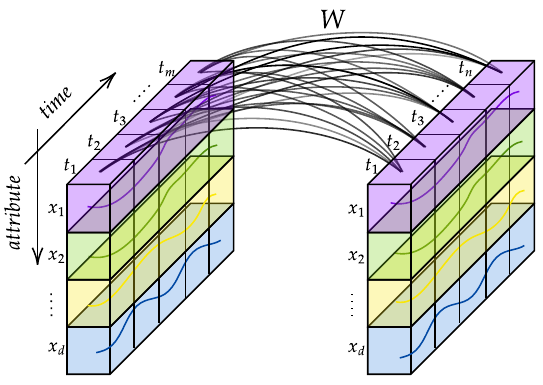}
    \caption{Visualizing the dense interpolation module. Where \(m\), \(n\), \(d\), are the length of the time series, the interpolation factor, and the number of time series attributes, respectively.
    In this work, we have that \(m=step_{in}\), \(d=d_{model}\) and \(n=M\).}
    \label{fig:denseinterp}
\end{figure}

\subsection{Dense Interpolation Layer}
\label{subsec:dil}
Unlike recurrent layers that produce a hidden state summarizing the information from the entire multivariate input sequence~\cite{hochreiter1997long}, the attention-based module returns a high-dimensional representation. 
To create a concise representation while capturing temporal structure and preserving temporal order, we introduce a dense interpolation layer (DIL) (aquamarine block in~\autoref{fig:architecture}). 
This layer is inspired by an interpolation algorithm originally developed for language modeling~\cite{trask2015modeling} and later applied to time-series data~\cite{song2018attend}.
The idea is to interpolate between the attention module's hidden representations  \(\textbf{s}_t\in\mathbb{R}^{d_{model}}\) at each time step \(t\in\{1,\dots,\tilde{T}\}\) with \(\tilde{T}\) being the total number of time steps in the input sequence, weighted by a function of their relative position in the final representation of length \(M\):
 
\begin{equation*}
	w_{t,m}=\left(1-\frac{\left|s_t-m\right|}{M}\right)^2, \forall \enskip t \in \{1,\dots,\tilde{T}\}  \enskip \textrm{and} \enskip m \in \{1, \dots, M\}
\end{equation*}

where \(s_t=M\cdot\frac{t}{\tilde{T}}\) represents the relative position of time step \(t\) in the final representation, and \(m\) is the index in the final representation.
This weight reflects the contribution of \(s_t\) to the final representation at position \(m\), decaying quadratically as the relative distance increases, ensuring the temporal structure is captured in the final output.
This interpolation is efficiently implemented performing the following matrix operation: \(\textbf{U}=\textbf{S}\times \textbf{W}\), where \(\textbf{S}=\left[\textbf{s}_1,\dots,\textbf{s}_{\tilde{T}}\right]\in \mathbb{R}^{d_{model}\times \tilde{T}}\) is the matrix containing the hidden states, \(\textbf{W}\in \mathbb{R}^{\tilde{T}\times M}\)  stores the interpolation weights \(w_{t,m}\), and \(\textbf{U}\in \mathbb{R}^{d_{model}\times M}\) is the final interpolated output representation.

We further enhance the DIL by allowing the weights \(w_{t,m}\) to be learned via the back-propagation process, enabling a more flexible and adaptive representation.
\autoref{fig:denseinterp} depicts this dense interpolation layer, where edge colors correspond to the magnitude of the learned weights, with darker edges representing stronger contributions and lighter edges indicating weaker ones. 
Finally, the unified hidden representation can be obtained by several methods, such as concatenation across all time steps. 
In this work, we use the last embedding as the final representation, summarizing the entire input sequence via the dense interpolation module.

\subsection{Multi-Level Joint Fusion}
The dense interpolation stage leads to the generation of three representation vectors in \(\mathbb{R}^{d_{model}} \), each representing the \ac{pv} production history, weather history, and weather forecast branches, respectively. 
To effectively combine these branch-specific unimodal representations into a unified multimodal vector, we use a fusion layer, also named shared representation layer.
Traditional fusion approaches, such as element-wise summation, mean, or Kronecker product, operate at the element-wise or tensor level, applying uniform operations to combine information across modalities. 
While these methods are effective for some tasks, 
they often do not fully account for the varying relevance of each input branch and fail to fully capture the intricate interactions and dependencies that exist between heterogeneous data sources, especially when the data streams represent temporally or spatially distinct phenomena~\cite{baltruvsaitis2018multimodal}, as in our case.
To address these limitations, we propose a soft-attention mechanism to guide the fusion process.
This attention mechanism helps emphasize the important features within each representation, contributing more effectively to the final multimodal vector. 
Soft-attention mechanisms have been widely adopted in various tasks, where selective focus has been shown to significantly improve model performance by enabling the model to prioritize informative features while ignoring less relevant ones~\cite{bahdanau2014neural, vaswani2017attention}.

Formally, let \(\textbf{x}_1, \textbf{x}_2 \in \mathbb{R}^{d_{\text{model}}}\) represent two input vectors from different modalities. 
Each input vector is passed through a fully connected layer followed by a non-linear activation function to obtain an intermediate representation:
\begin{equation*}
	 \textbf{h}_i = \tanh(\textbf{W}\textbf{x}_i + \textbf{b}), \quad i = 1, 2
\end{equation*}
where \(\textbf{W}\in \mathbb{R}^{d_{attn}\times d_{model}}\) and \(\textbf{b}\in \mathbb{R}^{d_{attn}}\) are learnable parameters.
To assess the relevance of each branch, an importance score \( s_i \) is computed as the inner product between the intermediate representation \( \mathbf{h}_i \) and a learnable context vector \( \mathbf{u} \in \mathbb{R}^{d_\text{attn}} \), capturing modality-specific weighting information:
\begin{equation*}
	s_i = \textbf{u}^T\textbf{h}_i, \quad i = 1, 2
\end{equation*}
The normalized importance weights \(\alpha_i\) for each branch are then obtained using the softmax function, ensuring that the  weights are non-negative and sum to 1, thus providing an interpretable probability distribution over the branches~\cite{bishop2006pattern, goodfellow2016deep}:
\begin{equation*}
	\alpha_i = \frac{\text{exp}\left(s_i\right)}{\sum_{k=1}^2{\text{exp}\left(s_k\right)}}, \quad i = 1, 2
\end{equation*}
where \(\alpha_i\) denotes the contribution of the \(i\)-th branch to the final representation. 
Finally, the output of the soft-attention mechanism, denoted by \(\textbf{x}_{attn}\), is computed as the weighted sum of the input vectors:
\begin{equation*}
	\textbf{x}_{attn} = \sum_{i=1}^2{\alpha_i \textbf{x}_i}
\end{equation*}
This formulation allows the model to dynamically adjust the contributions of each input modality, ensuring that the resulting multimodal representation captures both temporal and cross-modal dependencies effectively. 
Such an approach is particularly advantageous in our scenario, where the input branches represent different temporal aspects of the same phenomenon, including historical \ac{pv} production and weather data, as well as future weather forecasts. 
\bb

Given that the input modalities observe the phenomenon at distinct points in time, with the historical data capturing past events and the weather forecasts looking ahead, we implement a multi-level fusion approach to integrate information at varying levels of abstraction (black block in~\autoref{fig:architecture}).
At the first fusion level, the two temporally correlated sequences, \ac{pv} production history and weather history, are fused using the soft-attention mechanism previously described (orange block in~\autoref{fig:architecture}). 
The resulting fused vector is then processed through a fully connected module consisting of a dropout regularization layer~\cite{srivastava2014dropout} and a dense layer with ReLU as an activation function. 
At the second fusion level, this hidden intermediate representation is then fused with the weather forecast representation from the dense interpolation layer using another soft-attention mechanism (yellow block in~\autoref{fig:architecture}).
This joint representation is subsequently processed by a fully connected module comprising a dropout layer and a dense layer with ReLU activation, resulting in the unified multimodal representation denoted as:
\begin{equation}
    \textbf{x}_{joint}\in\mathbb{R}^{d_{model}}
\end{equation}
which serves as the input to the prediction head described in~\autoref{sec:pred}.
This hierarchical scheme allows the model to progressively integrate historical and forecast information while retaining the most informative features from each modality.

\subsection{Prediction Head and Physical Constraints}
\label{sec:pred}
The prediction head (red block in~\autoref{fig:architecture}) receives the multimodal representation \(\mathbf{x}_{\text{joint}}\) and maps it to the normalized PV production sequence.
This module consists of a dropout layer, a linear projection to match the forecast horizon \(w_{ahead}\), and a sigmoid activation function. 
Formally, the normalized prediction vector is expressed as:
\begin{equation*}
\hat{\mathbf{y}}=\sigma\left(Linear\left(Dropout\left(\mathbf{x}_{joint}\right)\right)\right), \quad \hat{\mathbf{y}}\in\left[0, 1\right]^{w_{ahead}}
\end{equation*}
where \(\sigma(\cdot)\) is the sigmoid function. 

This design inherently enforces two physical constraints: predictions are non-negative and upper-bounded by 1, consistent with the normalized PV time series introduced in~\autoref{subsec:ausgrid}.
In this way, the predicted values directly correspond to normalized specific yields. 
If needed, forecasts in physical units can be obtained by rescaling with the aggregated installed capacity:
\begin{equation*}
    \hat{\mathbf{P}} = \hat{\mathbf{y}} \cdot \sum_{i=1}^NkWp_i
\end{equation*}

\section{Experimental Setup and Results}
\label{sec:result}
In this section, we initially present the experimental setup describing the training details and the evaluation metrics adopted. 
Then, we introduce thirteen competitors, including statistical methods, methods specific to the Ausgrid dataset~\cite{fentis2020machine, kaur2021bayesian, kaur2023vae, kaur2022bayesian}, which are already presented in~\autoref{sec:relatedwork}, and deep neural networks.
Finally, we present and discuss the results.

\subsection{Training Details}
\label{subsec:training}
We use as initial dataset \ac{pv} power production and weather records from 1 July 2010 to 30 June 2012, splitting it into an 80-20 ratio to create the training and validation sets. 
The test set contains timestamps ranging from 1 July 2012 to 30 June 2013.

To ensure robustness, reproducibility, and fair comparison, no hyperparameter tuning was performed for MATNet, as hyperparameter optimization is highly dataset-specific and was considered out of the scope of this work. 
This choice is also consistent with the \emph{No Free Lunch} theorems for optimization, which state that no single algorithm or parameter configuration can be optimal across all possible problem domains~\cite{wolpert2002no}. 
In line with this theoretical result, empirical evidence has further shown that hyperparameter tuning does not necessarily yield substantial improvements over default configurations~\cite{arcuri2013parameter}. 
Unless otherwise specified, we use the following parameters for our architecture: input embedding dimensionality \(d_{model} = 512\), parallel heads attention layers \(h = 8\), number of sub-encoder-layers \(L=3\), the interpolation factor in the interpolation module \(M=w_{back}\), time steps in the input sequence \(w_{back}=24\), time steps in the output sequence \(w_{ahead}=24\).
We adopt a 24-hour time horizon because the day-ahead prediction of renewable generation is the most common requirement of smart grid management algorithms (\textit{e.g.} \cite{conteDA,zhang,conteDA2,michael}), further considering the daily cyclical nature of \ac{pv} generation.

Our proposed MATNet architecture and the entire training process are based on the PyTorch framework~\cite{paszke2019pytorch}. 
We trained all the models on a workstation with 1 NVIDIA A100 GPU for \(200\) epochs. 
The model weights and biases are updated using the Mean Squared Error (MSE) as a loss function, which measures the average squared difference between the predicted and the actual values.
We used the Adam optimizer~\cite{kingma2014adam} with a learning rate of \(10^{-3}\) and with \(\beta_1\) and \(\beta_2\) equal to \(0.9\) and \(0.999\), respectively. 
We reduced the learning rate throughout training by a factor of \(0.2\) once learning stagnates, and no improvement is seen for a patience number of epochs equal to \(20\).
The best-performing model on the validation set was used for final testing on the independent test set.
The code is publicly available on GitHub\footnote{The code is available at the following GitHub repository \url{https://github.com/cosbidev/MATNet}.}.

\subsection{Evaluation Metrics}
To assess the quality of the proposed forecast model, we employed standard evaluation metrics commonly used in forecasting~\cite{hyndman2006another}, providing different insights on the prediction errors.
In the following equations, \(n\) is the number of observations, \(y_i\) is the actual value of the \(i\)-th observation, and \(\hat{y}_i\) is the predicted value of the \(i\)-th observation:
\begin{itemize}
\item \textbf{Root Mean Squared Error}: it is calculated as the square root of the mean of the squared differences between the predicted and actual values: \bb
\begin{equation}
\label{eq:rmse}
        \text{RMSE} = \sqrt{\frac{1}{n}\sum_{t=1}^{n} {\left( y_t - \hat{y}_t \right)^2}}
\end{equation}
\item \textbf{Mean Absolute Error}: it measures the average absolute difference between the predicted and actual values:
\begin{equation}
\label{eq:mae}
    \text{MAE} = \frac{1}{n}\sum_{t=1}^{n} {\left| y_t - \hat{y}_t \right|}
\end{equation}
\item \textbf{Weighted Mean Absolute Percentage Error}: it measures the accuracy of a forecast, where the importance of the forecast errors is weighted according to the size of the actual values being forecast:
\begin{equation}
\label{eq:wmape}
    \text{wMAPE} = \frac{\sum_{t=1}^{n}{\left|y_t-\hat{y}_t\right|}}{\sum_{t=1}^{n}\left|y_t\right|}
\end{equation}
\item \textbf{Mean Absolute Scaled Error}: it measures the accuracy of a forecast, where the forecast errors are scaled by the mean absolute error of a na\"ive forecast:
\begin{equation}
\label{eq:mase}
    \text{MASE} = \frac{\frac{1}{n}\sum_{t=1}^{n}{\left|y_t-\hat{y}_t\right|}}{\frac{1}{n-1}\sum_{t=2}^{n}\left|y_t-y_{t-1}\right|}
\end{equation}

\end{itemize}
Each metric is in \(\mathbb{R}^{+}_{0}\), where lower values indicate more accurate forecasts.

\subsection{Competitors}
We compare the performance of MATNet against four groups of competitors. 

The first pool of competitors includes four well-established statistical time series forecasting models: AutoARIMA~\cite{hyndman2008automatic} and AutoCES~\cite{svetunkov2022complex}, both of which are statistical models that automatically select model parameters for each time series based on the Akaike information criterion~\cite{sakamoto1986akaike}; NPTS~\cite{fan2008nonlinear}, a non-parametric method based on local forecasting that assumes a non-parametric sampling distribution; and Theta~\cite{assimakopoulos2000theta}, a dynamic forecasting technique that is particularly effective for non-seasonal or deseasonalized time series.

Second, we included the four approaches~\cite{fentis2020machine,kaur2021bayesian,kaur2023vae,kaur2022bayesian} specific for the Ausgrid dataset, already introduced in~\autoref{sec:relatedwork}.

The third group of competitors includes several variations of our MATNet approach where we substitute the self-attention mechanism with four alternative recurrent-based backbones, which share the same multi-level fusion and final fully connected layers, whilst the input embedding, positional encoding, and dense interpolation modules are no longer necessary.
The purpose of this comparison is to assess the impact of self-attention on the model's performance relative to recurrent architectures. 
The four backbones we consider are: \ac{lstm}~\cite{hochreiter1997long}, bi-directional LSTM (BiLSTM)~\cite{graves2005framewise}, \ac{gru}~\cite{chung2014empirical}, and \ac{bigru}~\cite{dong2016character}. 

The final group of competitors includes two additional MATNet variants designed to evaluate the role of the DIL module. 
The first, \emph{cDIL-based MATNet}, replaces the DIL with its non-trainable counterpart originally proposed in~\cite{trask2015modeling}, as discussed in~\autoref{subsec:dil}. 
The second, \emph{MATNet w/o DIL}, removes the interpolation module and relies on last-token aggregation.

Note that we implemented all the methods except for the statistical models, which are provided by AutoGluon~\cite{shchur2023autogluon}, a Python AutoML framework for probabilistic time series forecasting. 
While AutoGluon supports probabilistic forecasts, we adapted these models for point forecasting by averaging their outputs.
An overview of the main configuration settings adopted for all comparative models is provided in~\refappendix{app:bas-config}.

\subsection{Results and Discussion}
\label{subsec:resultsdisc}
In this section, we comprehensively evaluate the performance of MATNet across multiple benchmark scenarios and analysis dimensions.
We begin by reporting results on the Ausgrid benchmark and proceed to analyze the impact of different input modalities through ablation studies.
We then assess the model’s sensitivity to missing data, followed by an evaluation of cross-site zero-shot generalization on five external PV datasets.
Finally, we examine the computational complexity of MATNet.

\subsubsection{Comparative Analysis}
\autoref{tb:metrics}  reports the performance in terms of RMSE, MAE, wMAPE, and MASE of our proposed model on the Ausgrid benchmark, followed by other thirteen rows presenting competitors' results, divided as described in the previous section.

Let us first examine MATNet's performance (first row in~\autoref{tb:metrics}) against the four statistical time series forecasting methods (rows 2-5). 
The results clearly indicate that MATNet significantly outperforms these statistical models across all metrics and all observed differences are statistically significant with a \(p\)-value \(< 0.001\), according to the Diebold-Mariano test~\cite{diebold2002comparing}, which is always used hereinafter.
These results are likely attributed to MATNet's ability to capture complex nonlinear relationships within the data, which traditional statistical methods struggle to model effectively. 
Statistical methods rely heavily on predefined assumptions about the data and fail to adapt to intricate temporal dependencies arising in PV power forecasting, particularly under varying weather conditions.

\begin{table}
\centering
\caption{Comparative analysis of the proposed method for \ac{pv} power production forecasting. In bold we highlight the best results for each metric.}
\label{tb:metrics}
\resizebox{\columnwidth}{!}{
\begin{tabular}{lllllll}
\toprule
\textbf{} & \textbf{Architecture}  & \textbf{RMSE \(\downarrow\)} & \textbf{MAE \(\downarrow\)} & \textbf{wMAPE \(\downarrow\)} & \textbf{MASE \(\downarrow\)} 
\\
\midrule
& MATNet           & \textbf{0.0445}        & \textbf{0.0243}       & \textbf{0.1672}         & \textbf{0.4072}  \\
\midrule

\multirow{4}{*}{\rotatebox[origin=c]{90}{\makecell{Statistical \\ based}}}  

& NPTS 
& 0.1450
& 0.0859
& 0.6483
& 1.6155
\\
& AutoARIMA 
& 0.1698
& 0.1017
& 0.6468
& 1.7075
\\
& AutoCES 
& 0.2014
& 0.1582
& 0.9847
& 2.6068
\\
& Theta 
& 0.1887
& 0.1530
& 1.0654
& 2.7107
\\
\midrule
\multirow{4}{*}{\rotatebox[origin=c]{90}{\makecell{Ausgrid \\ benchmark}}}  
& LsSVR~\cite{fentis2020machine}
& 0.1469  
& 0.0866 
& 0.7526 
& 1.7864 \\
& Bayes \ac{bilstm}~\cite{kaur2021bayesian} & 
0.1262  
 & 0.0840 
 & 0.6867
 & 1.6577 \\
& \(\alpha\)-\(\beta\) Bayes \ac{bilstm}~\cite{kaur2022bayesian}  & 0.1319 & 0.0856  & 
 0.7573 & 
 1.7892  \\ 
& VAE Bayes \ac{bilstm}~\cite{kaur2023vae}
 & 0.1317 
 & 0.0860
 & 0.7737 
 & 1.8189
\\
\midrule

\multirow{5}{*}{\rotatebox[origin=c]{90}{\makecell{MATNet \\ variants}}} 
& \ac{lstm}-based MATNet                                                   & 0.0517                                                & 0.0288                                                & 0.1993                                               & 0.4842                                                \\
& \ac{gru}-based MATNet                                                        & 0.0505                                                & 0.0284                                             & 0.1976                                               & 0.4824                                                \\
& \ac{bilstm}-based MATNet                                                      & 0.0510                                              & 0.0285                                                & 0.1949                                                & 0.4730                                              \\
& \ac{bigru}-based MATNet                                                        & 0.0484                                               & 0.0271                                                & 0.1864                                                & 0.4538                                               \\
& cDIL-based MATNet	& 0.1333	& 0.0812	& 0.7196	& 1.6933  \\
& MATNet w/o DIL & 0.0459 & 0.0255 & 0.1746 & 0.4249 \\
\bottomrule
\end{tabular}
}
\end{table}

\begin{figure*}
    \centering
\includegraphics[width=0.7\textwidth]{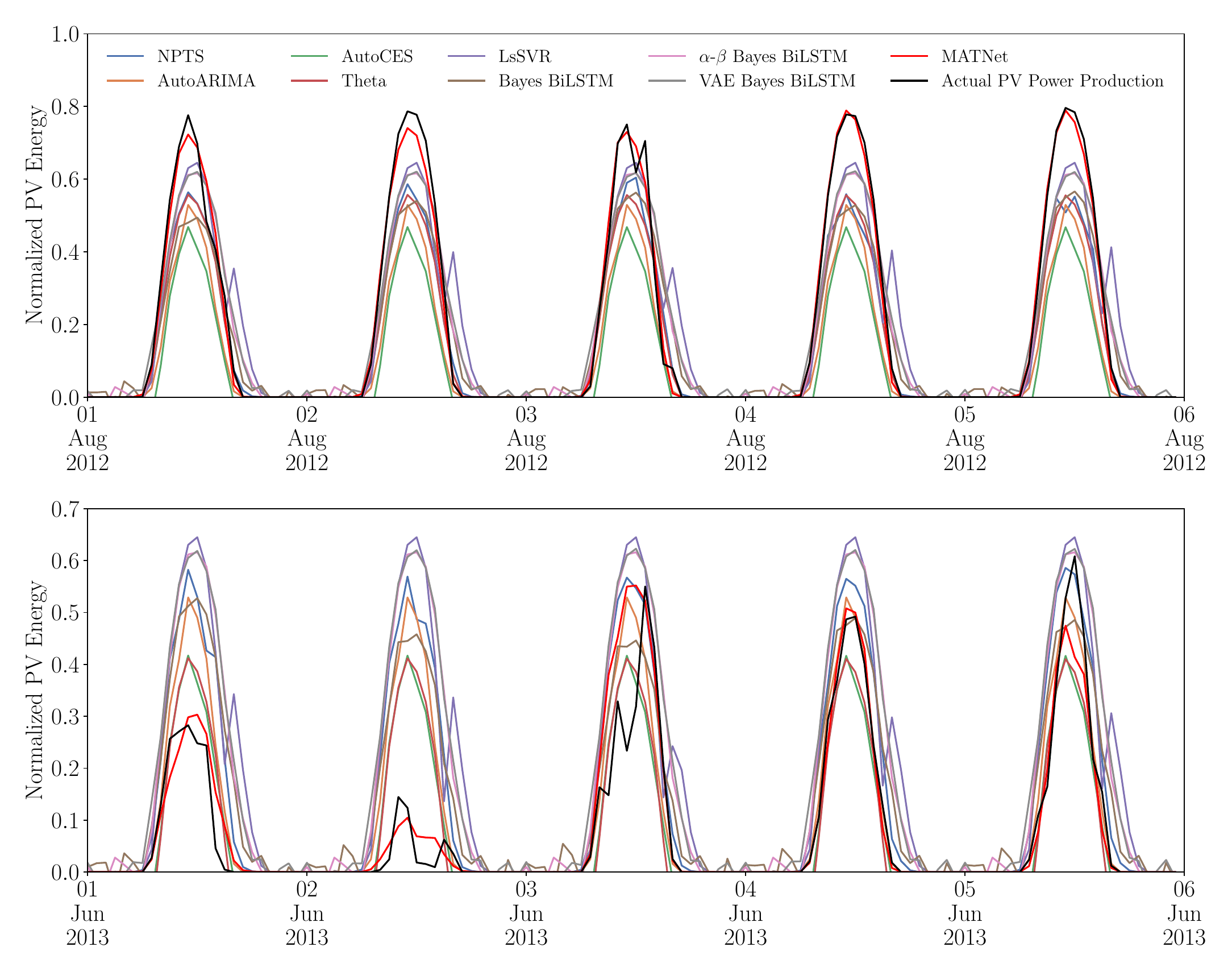}
    \caption{Comparison of hourly normalized PV power production (solid black line) with predictions from various models over two five-day periods (August 1-5, 2012 and June 1-5, 2013). 
    The normalized PV energy is plotted on the y-axis, while the x-axis represents the date. %\ggr cambia caption su y-axis come "Normalized energy yield" o "Normalized energy generation" \bb
    }
    \label{fig:comparisons}
\end{figure*}

Next, we compare MATNet against the four approaches (rows 6-9) representing the current state-of-the-art on the same benchmark~\cite{fentis2020machine, kaur2021bayesian, kaur2023vae, kaur2022bayesian}\footnote{It is worth noting that we re-implemented these methods to ensure consistency with our validation; therefore, the results in our study may differ from those originally reported.}: MATNet significantly outperforms these four state-of-the-art methods on all the evaluation metrics, with statistically significant differences at a \(p\)-value \(<0.001\). 
Since such four approaches are unimodal, we attribute MATNet's enhanced performance to its multimodal fusion that exploits interdependencies not captured by previous methods.

Turning our attention to the comparison with the four recurrent-based backbones (rows 10-13), we notice that our attention-based architecture outperforms all of them, and the differences are all statistically significant with a \(p\)-value \(< 0.01\). 
Here the key advantage of MATNet lies in its self-attention mechanism, which dynamically focuses on the most relevant features, rather than relying solely on sequential dependencies. 
This flexibility allows MATNet to capture short-term and long-term patterns in PV power production, which is crucial for accurate forecasting under fluctuating conditions.
Furthermore, it is worth noting that all four recurrent-based MATNet variants also statistically outperform the competitors included in the statistical and state-of-the-art approaches (\(p\)-values \(< 0.001\)). 
This fact highlights the effectiveness of the multi-level soft-attention multimodal strategy, which introduces substantial improvements in forecasting accuracy, regardless of the specific backbone used.
Finally, we evaluate the performance of the last two MATNet variants (rows 14–15), both designed to assess the role of the dense interpolation layer. MATNet significantly outperforms these variants, with differences statistically significant at \(p\)-value \(< 0.001\) and \(p\)-value \(< 0.01\), respectively, highlighting the contribution of the DIL to forecasting accuracy. 
The cDIL-based variant performs comparably to the statistical and state-of-the-art baselines, likely due to the limited adaptability introduced by its fixed-weight interpolation. 
In contrast, the MATNet w/o DIL variant achieves performance levels similar to the recurrent-based alternatives, suggesting that although its removal reduces accuracy, the architecture still benefits from multimodal design and attention-based fusion.

\autoref{fig:comparisons} illustrates a comparison of hourly normalized \ac{pv} power production against model predictions over two distinct five-day periods, specifically August 01-05, 2012, and June 01-05, 2013. 
The former, characterized by predominantly clear skies with occasional clouds and scattered rain, represents a relatively stable weather scenario. 
In contrast, the latter period, dominated by overcast skies and frequent rainfall, provides a more challenging test bed, enabling a comprehensive evaluation of model robustness under adverse weather conditions.
For the sake of visualization, we show only the statistics-based and state-of-the-art approaches, excluding MATNet variants with recurrent-based backbones, as previous paragraphs have already established the superiority of MATNet's architecture over its other variants.
In the first period reported in the top panel of~\autoref{fig:comparisons}, MATNet demonstrates superior performance in consistently tracking the actual PV power production compared to other competitors.
When examining the second period (bottom panel of~\autoref{fig:comparisons}), MATNet still exhibits a smaller deviation from the actual values relative to other models. 
This is especially pronounced during rapid fluctuations in power production caused by dynamic weather patterns. 
MATNet’s responsiveness and accuracy during these abrupt changes highlight its robustness, underscoring the efficacy of the attention mechanisms and multi-level joint fusion integrated into the forecasting process.
Despite these advantages, it is worth noting that the predictions are not flawless, particularly on June 3 and 5, 2013, where MATNet exhibits some difficulty in closely following the actual power trends.
This may be due to two key factors: first, the discrepancy between point-based weather measurements and the aggregate nature of the PV power production predictions, given that we consider 26 PV units distributed across an area of 75 square kilometers.
We hypothesize that predicting individual PV unit outputs using point-specific weather data could improve the model’s performance. 
Second, the need for a larger dataset to enhance the model's generalization capabilities.

\begin{figure*}
    \centering
    \includegraphics[width=0.8\textwidth]{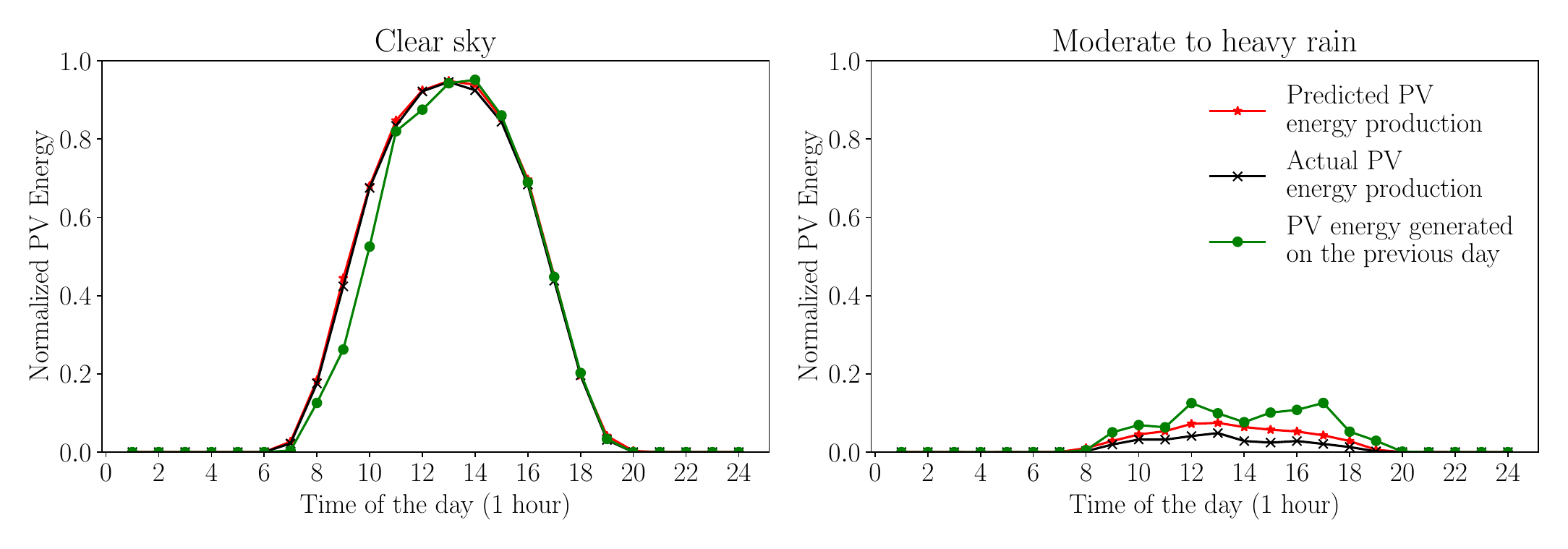}
    \caption{Forecasting results of MATNet on Ausgrid test set.
    On the left is the forecast for the best-performing day (2013-06-11), while on the right is the forecast for the worst-performing day (2013-01-28). 
    Both predictions are evaluated using the MASE metric. 
    %\ggr cambia caption su y-axis come "Normalized energy yield" o "Normalized energy generation" \bb
}
    \label{fig:predictions}
\end{figure*}

For the sake of completeness,~ \autoref{fig:predictions} shows the \ac{pv} production forecasts of MATNet for the best-performing (left) and worst-performing (right) day according to the MASE metric.
In black is the actual production curve and in red is the \ac{pv} production forecast curve; besides, to provide additional context, we have also included the \ac{pv} generation from the previous day, represented by the green curve.
The plot on the left reveals the effectiveness of our MATNet on its best-performing day, with a MASE score equal to \(0.0613\).
Predominantly clear skies characterize this day, and the forecast almost perfectly follows the actual \ac{pv} production. 
On the other hand, the graph on the right shows the worst-performing day with a MASE equal to \(2.2028\).
This day corresponds to a weather regime dominated by \emph{heavy-intensity rain}, observed in only 0.15\% of the training timestamps and thus extremely rare in the training distribution. 
The rarity of such adverse-weather events limits the ability of the attention mechanism to learn robust cross-modal dependencies, resulting in systematic under- or over-scaling of the predicted magnitudes despite the preservation of the diurnal trend. 
Extending the training period to additional years would broaden weather coverage and increase exposure to rare patterns, mitigating this limitation.
Even under these challenging conditions, the forecast qualitatively reproduces the temporal evolution of PV production, highlighting the resilience of MATNet, which continues to outperform state-of-the-art methods that typically experience a decline in performance during adverse weather situations.
The difference between the red (forecast) and green (previous day's production) patterns also highlights that the forecast is not solely based on the previous day's production, demonstrating the positive impact of incorporating weather forecasts in our model.

\begin{figure}[t]
    \centering
    \includegraphics[width=0.5\textwidth]{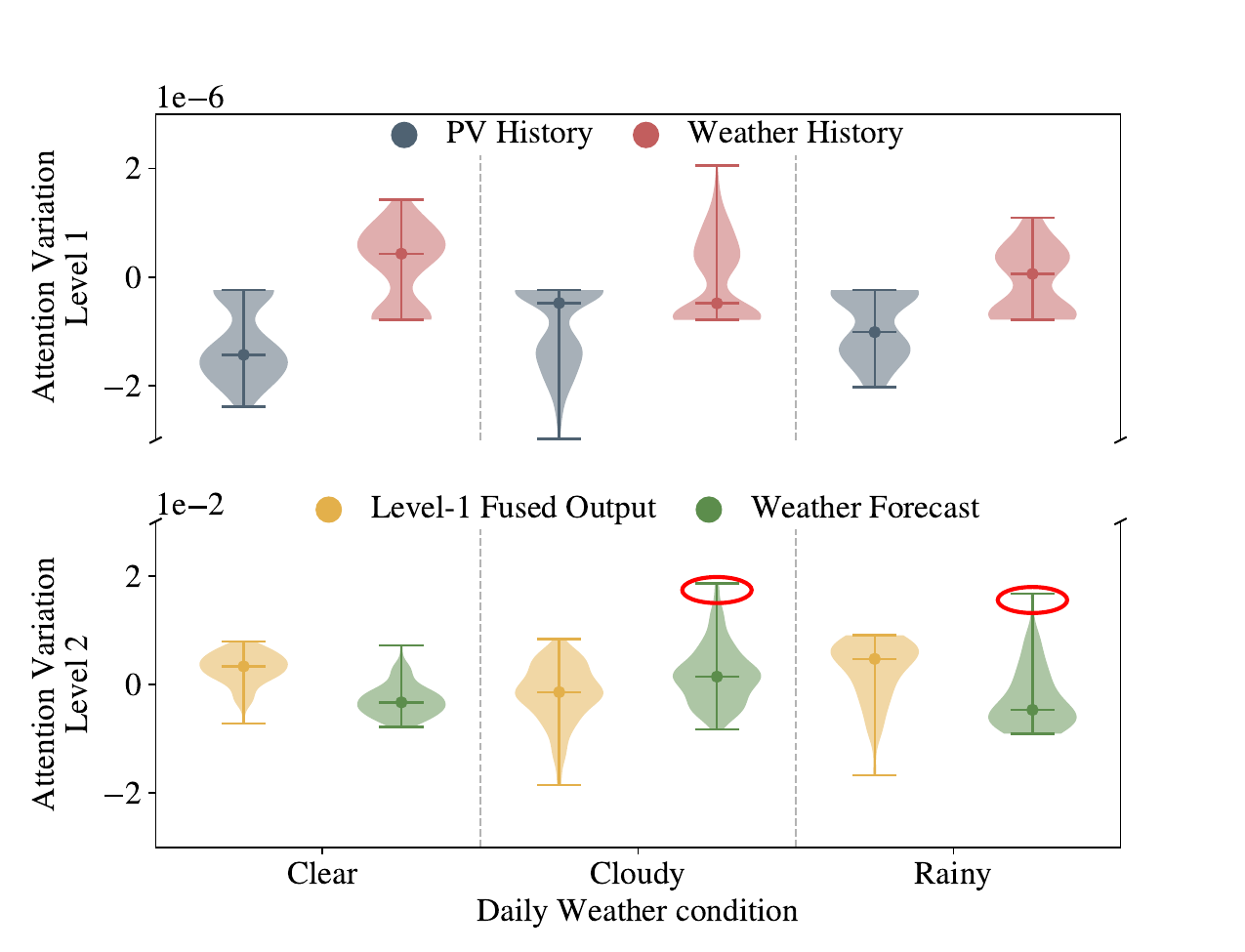}
    \caption{Violin plots of attention weight variations in the two-stage soft-attention fusion. 
    Top: first-level fusion between historical photovoltaic (\textcolor{pv}{PV}) and meteorological (\textcolor{weath}{Weather}) inputs. 
    Bottom: second-level fusion where the \textcolor{fused}{Level-1 fused output} is further combined with the \textcolor{forecast}{Weather Forecast}. 
    Each pair of violins is grouped by daily weather condition (\emph{Clear}, \emph{Cloudy}, and \emph{Rainy}).}
    \label{fig:attn_fusion}
\end{figure}

\paragraph{Analysis of Multi-Level Fusion Attention}
To further elucidate the interpretability of the proposed multi-level fusion mechanism and to reinforce the analysis of the worst-performing day,~\autoref{fig:attn_fusion} illustrates the distributions of attention weight variations learned by MATNet across the two fusion stages.
The violin plots are grouped by daily weather condition (Clear, Cloudy, and Rainy), showing how the modality-specific attention weights evolve under different meteorological regimes.

In the first fusion level (top panel), which combines PV History and Weather History modalities, the variations are on the order of \(10^{-6}\), indicating extremely stable attention behavior across samples.
In absolute terms, the attention weights averaged across samples are consistently dominated by PV History (\(\approx0.99\)), with Weather History contributing only marginally (\(\approx 0.01\)).
Such quasi-stationary behavior demonstrates that MATNet primarily relies on historical PV information to infer short-term generation trends, regardless of the prevailing meteorological conditions.
In contrast, in the second fusion level (bottom panel), where the \emph{Level-1 fused output} is integrated with the Weather Forecast branch, the variations increase to the order of 
\(10^{-2}\), revealing a stronger adaptive behavior.
Under cloudy and rainy regimes, the distributions become broader and more asymmetric, as the model dynamically reweights modality contributions in response to forecast uncertainty. In particular, the regions of higher dispersion correspond to increased attention toward the Weather Forecast modality (see red circles in~\autoref{fig:attn_fusion}), confirming that MATNet relies more heavily on predictive meteorological cues when PV dynamics become less deterministic.

These findings confirm the hierarchical role of MATNet’s multi-level soft-attention fusion: the first stage ensures stability by consistently leveraging historical PV information, whereas the second stage introduces adaptivity and cross-modal flexibility, enabling the model to maintain robustness and accuracy across diverse meteorological regimes.

\begin{table}
\centering
\caption{Ablation study results from the Ausgrid test set comparing the MATNet model's performance with various input branches disabled. 
In bold we highlight the best results per metric.}
\label{tb:ablation}
\resizebox{\columnwidth}{!}{
\begin{tabular}{cccccccc}
\toprule
\bfseries\makecell{\ac{pv} \\ Production} & \bfseries\makecell{Weather \\History} & \bfseries\makecell{Weather \\Forecast}  & \bfseries\makecell{RMSE \(\downarrow\)} & \bfseries\makecell{MAE \(\downarrow\)} & \bfseries\makecell{wMAPE \(\downarrow\)} & \bfseries\makecell{MASE \(\downarrow\)} \\
\midrule
\cmark  & \xmark & \xmark 
& 0.1078  
& 0.0629    
& 0.5689    
& 1.3433   
\\
\xmark & \cmark  & \xmark               
& 0.1381
& 0.0863
& 0.7508	
& 1.7756       
\\
\xmark & \xmark  & \cmark         
& 0.0815	
& 0.0547	
& 0.3986
& 0.9834       
\\
\cmark   & \cmark  & \xmark             
& 0.1024        
& 0.0592     
& 0.4972        
& 1.1916      
\\
\cmark  & \xmark & \cmark                
& 0.0455       
& 0.0247     
& 0.1714        
& 0.4159       
\\
\xmark  & \cmark  & \cmark           	
& 0.0454
& 0.0249	
& 0.1678	
& 0.4091        
\\
%\rowcolor{Gray}
\cmark & \cmark & \cmark              
& \textbf{0.0445}        & \textbf{0.0243}       & \textbf{0.1672}         & \textbf{0.4072}  \\
\bottomrule
\end{tabular}
}
\end{table}

\subsubsection{Ablation Analysis}
To deepen the results, and given that MATNet is a multimodal architecture incorporating heterogeneous input sources (i.e., \ac{pv} production data, weather history, and weather forecasts), we performed six ablation studies by systematically excluding one or two of these input modalities. 
In other words, for each study, we turned off the input branch corresponding to the removed data source,
allowing us to evaluate how the model would perform without it. 
The results of this analysis provide valuable insights into each input modality's importance, helping us to assess the effectiveness of MATNet and its robustness in scenarios with missing modalities.
The results are summarized in~\autoref{tb:ablation}, where the first three columns indicate the input modalities used in each experiment. 
Straightforwardly, the last row corresponds to the full MATNet configuration with all modalities enabled.

Focusing on the unimodal configurations (rows 1-3), we observe that PV production data and weather forecasts outperform all the statistical and state-of-the-art competitors across all metrics, with statistically significant differences (p-value \(< 0.01\)). 
The weather history modality also outperforms statistical models but performs comparably to the state-of-the-art methods. This outcome may be attributed to the nature of the historical weather data, which, while informative, lacks the predictive power of weather forecasts or PV data and may fail to capture rapid changes or anomalies effectively.
Among the unimodal configurations, the best performance is achieved when using only the weather forecast data. This result underscores the pivotal role of forecasted weather conditions in determining the model's success.

Shifting our attention to the bimodal configurations (rows 4-6), we observe that all bimodal combinations outperform the competitors across all metrics. 
These results further reinforce the notion that incorporating multiple heterogeneous data sources substantially enhances the model’s performance.
Notably, configurations that include the weather forecast modality (rows 5 and 6) consistently show significant improvements over all the statistical and state-of-the-art competitors, with \(p\)-values \(< 0.001\), and their performance is comparable to the full MATNet model, with no statistically significant differences (\(p\)-values \(> 0.05\)).

These findings suggest that MATNet maintains strong performance, even when one or more input modalities are unavailable, except for the unimodal configuration involving only the weather history modality. 
This robustness ensures that, in practical applications where data may be incomplete or unavailable, MATNet continues to function effectively and reliably in most scenarios.

\begin{table}[]
\caption{External validation sites with location, period, Köppen-Geiger climate type (BSk: cold semi-arid, Csa: Mediterranean hot summer, Cwa: humid subtropical with dry winter, Csb: Mediterranean warm summer, Cfa: humid subtropical with no dry season), and PV capacity (kWp).}
    \label{tab:ext_val}
    \centering
    \resizebox{\columnwidth}{!}{
    \begin{tabular}{cccccccc}
    \toprule
    \bfseries{Dataset} &
    \bfseries{Location}	& \bfseries{Latitude} & \bfseries{Longitude} & \bfseries{From date}	& \bfseries{To date} & \bfseries{Köppen-Geiger} &\bfseries\makecell{Installed Power \\ (kWp)} \\
    \midrule
    ~\cite{yao2021photovoltaic}
    & Hebei & 38.047 & 114.951 & 08/16/18  & 06/13/19 & BSk & 6600 \\
    ~\cite{sarmas2025photovoltaic} & Lisbon & 38.728 & -9.138 & 01/01/19 & 12/31/19 & Csa & 46 \\
    ~\cite{lin2024highresolution}
    & Hong Kong  & 22.340  & 114.260  & 
       08/24/21  & 08/24/22 & Cwa & 16.22 \\
       ~\cite{nie2023skipp} & Stanford & 37.427 & -122.174  & 01/01/18 & 12/31/18 & 
       Csb & 30 \\
       ~\cite{pecanstreet2009dataport}
       & Austin & 30.267 & -97.743 & 01/01/18 & 12/31/18 & Cfa & 6.3 \\
         \bottomrule
    \end{tabular}}
\end{table}

\begin{table*}
\caption{Performance comparison of MATNet and its backbone variants across five external benchmark datasets. 
Results are reported as RMSE and MAE (lower is better). 
Bold values indicate the best performance per metric and dataset; AvgWins denotes the percentage of best results for each method.}
    \label{tab:ext_val_res}
\centering
\resizebox{\textwidth}{!}{
\begin{tabular}{ccc|cc|cc|cc|cc|cc|cc}
\toprule
{\textbf{Methods}} & \multicolumn{2}{c}{\bfseries{MATNet}} & \multicolumn{2}{c}{\bfseries{LSTM-based MATNet}} & \multicolumn{2}{c}{\bfseries{GRU-based MATNet}} & \multicolumn{2}{c}{\bfseries{BiGRU-based MATNet}}  & \multicolumn{2}{c}{\bfseries{BiLSTM-based MATNet}}  & \multicolumn{2}{c}{\bfseries{cDIL-based MATNet}} & \multicolumn{2}{c}{\bfseries{MATNet w/o DIL}}\\
 
\textbf{Metric} & RMSE \(\downarrow\) & \multicolumn{1}{c|}{MAE \(\downarrow\)}   & RMSE \(\downarrow\) & \multicolumn{1}{c|}{MAE \(\downarrow\)} & RMSE \(\downarrow\) & \multicolumn{1}{c|}{MAE \(\downarrow\)}
& RMSE \(\downarrow\) & \multicolumn{1}{c|}{MAE \(\downarrow\)}
& RMSE \(\downarrow\) & \multicolumn{1}{c}{MAE \(\downarrow\)}
& RMSE \(\downarrow\) & \multicolumn{1}{c}{MAE \(\downarrow\)} & RMSE \(\downarrow\) & \multicolumn{1}{c}{MAE \(\downarrow\)}
\\
\midrule
Hebei
     & \textbf{0.0763} & \textbf{0.0438} & 0.0853 & 0.0495 & 0.0857 & 0.0486 & 0.0861 & 0.0492 & 0.0787 & 0.0454 & 0.1212 & 0.0763 & 0.0785 & 0.0445 \\
     Lisbon
     & \textbf{0.1768} & \textbf{0.1099} & 0.1847 & 0.1153 & 0.2146 & 0.1352 & 0.1942 & 0.1216 & 0.1835 & 0.1154 & 0.2284 & 0.1446 & 0.1848 & 0.1141\\
     Hong Kong
    & 0.1089 & \textbf{0.0610} & \textbf{0.1071} & \textbf{0.0610} & 0.1262 & 0.0726 & 0.1140 & 0.0654 & 0.1173 & 0.0675 & 0.1786 & 0.1076  & 0.1088 & 0.0611 \\
     Stanford
     & 0.0668 & 0.0398 & 0.0710 & 0.0426 & 0.0688 & 0.0413 & 0.0680 & 0.0403 & 0.0693 & 0.0412 & 0.1229 & 0.0792  & \textbf{0.0613} & \textbf{0.0360}  \\
    Austin
     & \textbf{0.0832} & \textbf{0.0481} & 0.0912 & 0.0529 & 0.0998 & 0.0588 & 0.0935 & 0.0550 & 0.0923 & 0.0542 & 0.1593 & 0.0961 & 0.0847 & 0.0487 \\
\midrule
AvgWins & \multicolumn{2}{c}{65\%} & \multicolumn{2}{c}{15\%} & \multicolumn{2}{c}{0\%} & \multicolumn{2}{c}{0\%} & \multicolumn{2}{c}{0\%} & \multicolumn{2}{c}{0\%} & \multicolumn{2}{c}{20\%} \\
\bottomrule
\end{tabular}
}
\end{table*}

\begin{figure}
    \centering
    \includegraphics[width=\linewidth]{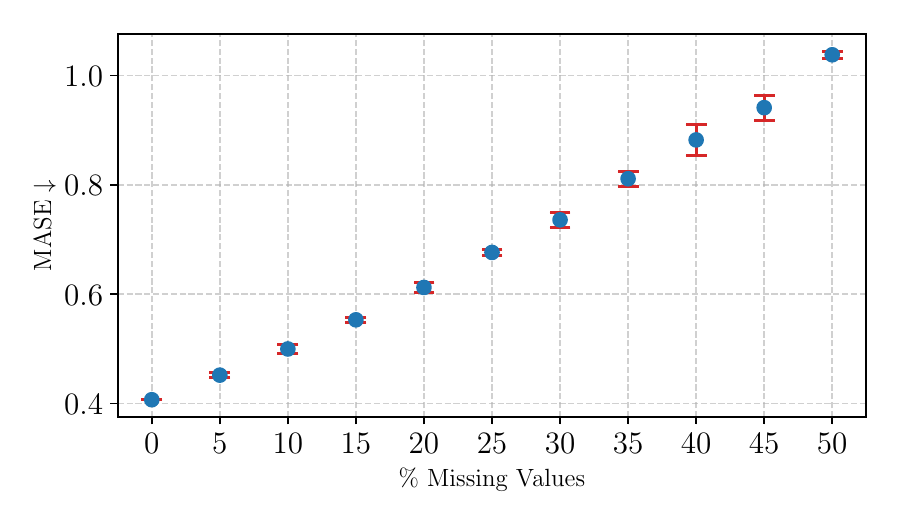}
    \caption{Evaluation of MATNet’s robustness to missing input data. Each point reports the MASE on the Ausgrid test set when a given percentage of input features is randomly masked (from 5\% to 50\%, in steps of 5\%) across all input modalities.}
    \label{fig:miss_data}
\end{figure}

\subsubsection{Sensitivity Analysis on Missing Data}
We evaluated MATNet’s robustness under conditions of incomplete input data, a common challenge in real-world deployments due to sensor faults, transmission errors, or data acquisition delays. 
To simulate such scenarios, we applied value masking during inference by randomly setting a growing proportion of input features to zero, ranging from 5\% to 50\% in increments of 5\%.
~\autoref{fig:miss_data} reports the MASE on the Ausgrid test set as a function of the proportion of masked inputs. 
MATNet maintains relatively stable performance up to approximately 20\% missing data, with only a modest increase in error. 
Beyond this threshold, performance degrades progressively. 
Around 35\% missingness leads to a doubling of MASE relative to the fully observed baseline, yet without abrupt failure, highlighting the model’s ability to degrade gracefully under increasing input corruption.
It is worth noting that even with 50\% of the input features missing, MATNet still outperforms all external baseline models operating under complete data availability (\autoref{tb:metrics}), including both statistical methods and Ausgrid-specific deep learning approaches. 
We attribute MATNet's resilience to its multimodal design: unlike competing models, which are univariate and limited in their ability to capture complex environmental dependencies, MATNet integrates multiple data sources, enabling it to infer missing information through cross-modal relationships. 
%This design supports more robust generalization in scenarios where partial data loss is unavoidable.

\subsubsection{Cross-Site Zero-Shot Generalization}
To further assess the generalizability and robustness of MATNet, we conducted a cross-site zero-shot generalization experiment, testing the model on five publicly available PV generation datasets, whose characteristics are summarized in~\autoref{tab:ext_val}.
These datasets differ from Ausgrid in multiple aspects. 
First, they vary significantly in terms of installed PV capacity, ranging from large-scale facilities (e.g., Hebei with 6600 kWp) to small residential setups (e.g., Austin with 6.3 kWp), whereas Ausgrid represents an aggregated signal from 26 households with moderate capacity. 
Second, the datasets are geographically diverse, covering four continents and spanning different hemispheres, which results in distinct seasonal patterns. 
Third, the underlying climate conditions are highly heterogeneous, ranging from semi-arid (BSk) to humid subtropical (Cfa), in contrast to the Cfa climate zone of the Ausgrid dataset. 
These substantial differences ensure that each external dataset represents a distinct operational regime, allowing us to assess the zero-shot generalization ability of MATNet under varying environmental, technical, and climatic conditions.
% The consistent performance observed across these domains supports the model’s robustness to unseen sites.

~\autoref{tab:ext_val_res} reports the results of the external validation in terms of RMSE and MAE for MATNet and its backbone variants across the five sites. 
The best-performing value for each metric and dataset is highlighted in bold, while the AvgWins row indicates the percentage of cases in which each method achieves the best result.
MATNet exhibits strong zero-shot generalization, consistently achieving the lowest errors on the majority of external sites, attaining the best performance in 65\% of cases. 
The model maintains high predictive accuracy despite substantial differences in site characteristics, climate patterns, and PV capacity. 
In comparison, recurrent-based variants, such as the LSTM-based MATNet, achieve competitive results on a single dataset (e.g., 15\% of best scores) but exhibit inferior and less consistent performance across datasets. 
The MATNet w/o DIL variant wins in 20\% of cases, confirming that removing the interpolation layer leads to a performance drop, thus validating the importance of temporal densification.
Notably, the cDIL-based MATNet variant exhibits the highest error rates across all sites, underscoring the critical role of the trainable dense interpolation layer for effective temporal modeling in diverse scenarios.

These results underscore the robustness and transferability of MATNet, demonstrating its capacity to generalize effectively to previously unseen and heterogeneous PV generation environments without requiring site-specific adaptation. The cross-site evaluation further highlights MATNet’s potential for practical deployment in real-world energy management scenarios across diverse operational contexts.

\begin{figure}
    \centering
    \includegraphics[width=\linewidth]{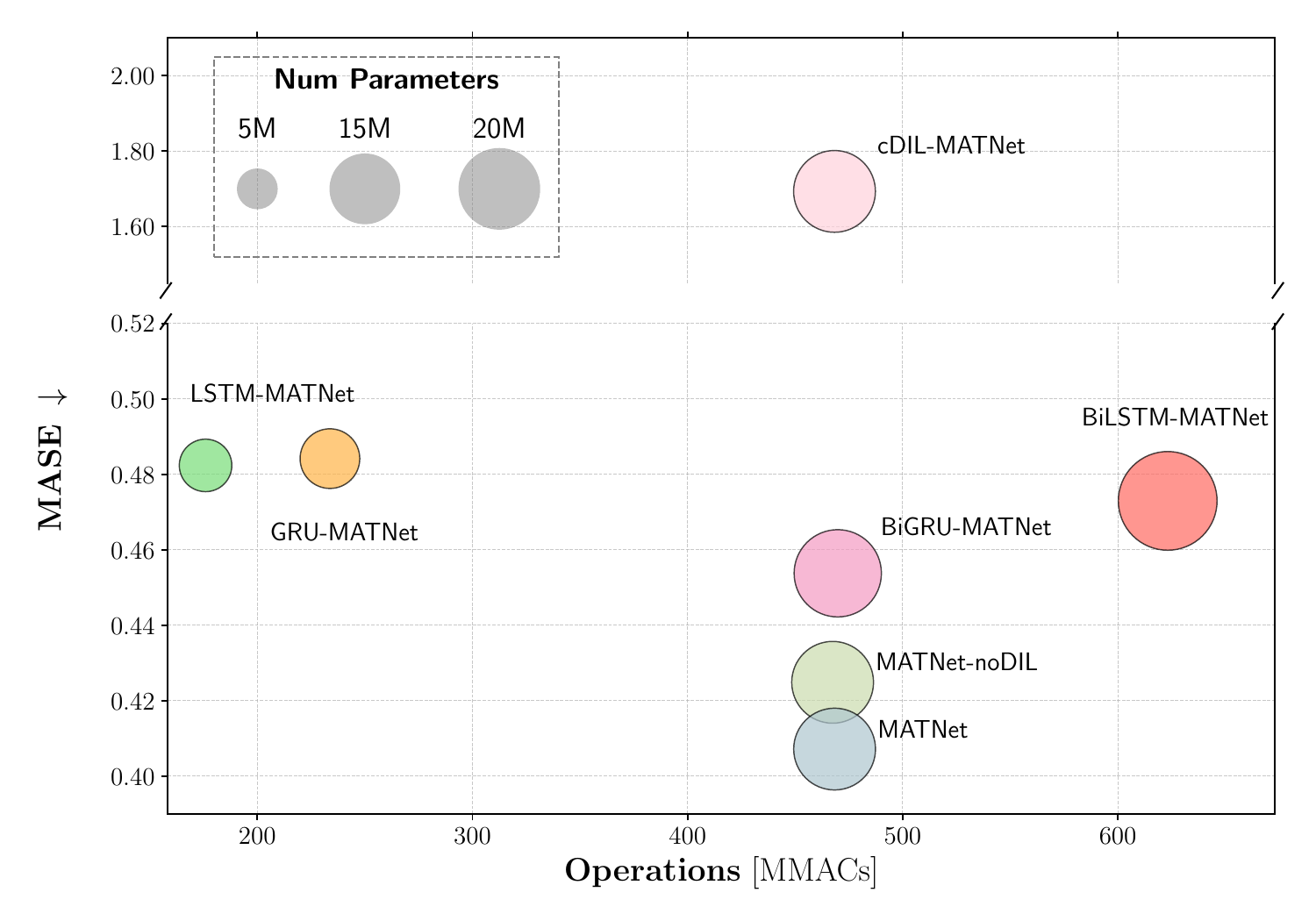}
    \caption{Comparison of MATNet and its backbone variants in terms of MASE (y-axis) and computational cost in MMACs (x-axis).
    Bubble size reflects the number of parameters, as shown by the reference bubbles in the top left (5M, 15M, and 20M parameters).}
    \label{fig:compl_anal}
\end{figure}

\subsubsection{Computational Complexity Analysis}
In practical forecasting applications, computational complexity is a key consideration alongside predictive accuracy.
\autoref{fig:compl_anal} illustrates the trade-off between computational cost and forecasting accuracy for MATNet and its backbone variants, summarizing the number of trainable parameters, inference-time MMACs, and test accuracy in terms of MASE.

MATNet achieves the lowest MASE with a moderate computational cost, making it the most efficient architecture among those considered. 
Recurrent-based models, such as BiLSTM-MATNet and BiGRU-MATNet, require significantly more operations and parameters while delivering inferior accuracy. 
The cDIL-based MATNet, in contrast, exhibits both higher computational cost and the highest MASE among all variants.
Traditional recurrent architectures like LSTM-MATNet and GRU-MATNet offer lower computational demands but do not achieve the same accuracy as MATNet. 
This indicates that, despite their efficiency, they are less suitable for high-fidelity PV forecasting in diverse scenarios. 
MATNet w/o DIL achieves a slightly lower computational cost than the full MATNet. However, this minor gain in efficiency comes with a drop in accuracy, underscoring the contribution of the DIL module to MATNet’s superior forecasting performance. 
These results confirm that MATNet provides the best balance between accuracy and computational efficiency, supporting its deployment in real-world energy forecasting applications where both performance and resource constraints must be considered.

\section{Conclusion}
\label{sec:conclusion}
In this work, we proposed MATNet, a novel multimodal self-attention architecture for multi-step, day-ahead \ac{pv} power generation forecasting. 
The architecture leverages historical \ac{pv} production data, historical weather data, and weather forecasts as inputs, and integrates these heterogeneous modalities through a multi-level soft-attention fusion strategy that effectively combines information at different levels of abstraction.

We extensively evaluated MATNet on the Ausgrid benchmark dataset, where it significantly outperformed thirteen baseline models, including statistical, machine learning, and deep learning approaches.
Specifically, MATNet achieved a RMSE of 0.0445 and a MASE of 0.4072, corresponding to relative improvements of approximately 65\% and 75\%, respectively, compared to the best-performing external baseline (Bayes BiLSTM).
The model’s ability to capture temporal and cross-modal dependencies through its soft-attention fusion framework was key to achieving this superior predictive performance.
We further assessed MATNet’s robustness to incomplete inputs through a sensitivity analysis on missing data. 
MATNet maintained stable performance up to 20\% missingness and showed graceful degradation under higher corruption levels. 
Even under 50\% missingness, the model outperformed all external baselines, confirming its ability to recover informative signals via cross-modal inference.
Furthermore, MATNet demonstrated strong generalization capabilities through a cross-site zero-shot evaluation conducted on five external PV datasets with diverse operational contexts, climate types, and installed capacities.
The model consistently achieved high forecasting accuracy under substantial domain shifts without requiring site-specific adaptation.
In addition, our computational complexity analysis confirmed that MATNet provides an effective balance between predictive accuracy and resource efficiency, making it suitable for practical deployment in real-world forecasting applications where both performance and computational cost are critical considerations.

While the proposed method achieves remarkable forecasting accuracy, it relies on the integration of forecast and historical weather data to enhance \ac{pv} power production predictions.
It is worth recognizing that the acquisition of these data can involve costs in some cases.
However, the services utilized in our research provide access to these data free of charge for household \ac{pv} systems. 
Therefore, incorporating meteorological data does not impose any financial burden for the specific use case of household applications.
We also note that, due to the lack of available historical forecast data for the study period, our model relies on a simulated forecast input derived from past weather observations. 
While this strategy introduces realistic sources of uncertainty, we acknowledge that it cannot fully replicate the behavior of real operational forecasts.
%\hl{Nonetheless, we acknowledge that the simulation of forecast inputs based on historical weather data, while realistic in nature, does not fully replicate the behavior of real operational forecasts.}
Moreover, the dataset used in this study was limited to 26 residential PV units with complete and high-quality data. While this selection ensured consistent and reliable measurements, it may also introduce a mild selection bias toward more stable systems, potentially reducing the model’s generalizability to larger or noisier datasets. Future work will address this limitation by incorporating data imputation strategies and by extending the training to larger, more diverse PV datasets to further test the scalability and robustness of the proposed approach.

To further improve the proposed method by making it more resilient, robust, and reliable, future work is directed toward developing an adaptive joint fusion layer.
The idea is to dynamically weight the contribution of the various input modalities to provide more weight to the weather forecast branch during adverse conditions.
In this way, the model can better adjust its predictions to follow the typical silhouette of \ac{pv} generation in clear weather, while giving more weight to the branch of the weather forecast in case of adverse conditions.
Besides, another worthy future direction is the incorporation of the eXplainable \ac{ai} paradigm into the proposed framework. 
By making the proposed framework interpretable, we can facilitate the integration of \ac{ai}-based \ac{res} into the power systems, enabling energy stakeholders to make informed, evidence-based decisions.
In this context, investigating feature importance and selection strategies could enhance interpretability and reveal the contribution of individual weather variables to the model's predictions, offering additional insights for decision-makers.

\section*{CRediT authorship contribution statement}
\textbf{Matteo Tortora}: Conceptualization, Methodology, Software, Validation, Formal analysis, Investigation, Resources, Data Curation, Writing - Original Draft, Writing - Review \& Editing, Visualization.
\textbf{Francesco Conte}: Conceptualization, Funding acquisition, Writing - Review \& Editing. \textbf{Gianluca Natrella}: Data Curation.
\textbf{Paolo Soda}: Conceptualization, Methodology, Funding acquisition, Investigation, Writing - Review \& Editing.

\section*{Declaration of competing interest}
All authors declare that they have no known competing financial interests or personal relationships that could have appeared to influence (bias) the work reported in this paper.

\section*{Acknowledgments}
This work was partially supported by (i) PNRR MUR project PE0000013-FAIR; (ii) Project ECS 0000024 Rome Technopole, - CUP C83C22000510001,  NRP Mission 4 Component 2 Investment 1.5,  Funded by the European Union - NextGenerationEU; (iii) the National Recovery and Resilience Plan (NRRP), Mission 4 Component 2 Investment 1.3 - Call for tender No. 1561 of 11.10.2022 of Ministero dell’Università e della Ricerca (MUR) in NEST SPOKE 8.

\appendix
\section{Selected Ausgrid customers}
\label{app:ausgridcustomers}
In this work, we only focused on 26 out of the 300 customers: 33, 47, 73, 87, 88, 110, 124, 144, 151, 153, 157, 163, 175, 176, 188, 200, 201, 207, 222, 240, 256, 259, 263, 272, 281, 293, for a total of 8 postal codes in Newcastle region. These have no missing values or artifacts due to inactivity, failures, or interruptions.

\section{Sensitivity Analysis on Input Granularity}
\label{app:granular}
\begin{figure}
    \centering
    \includegraphics[width=0.5\textwidth]{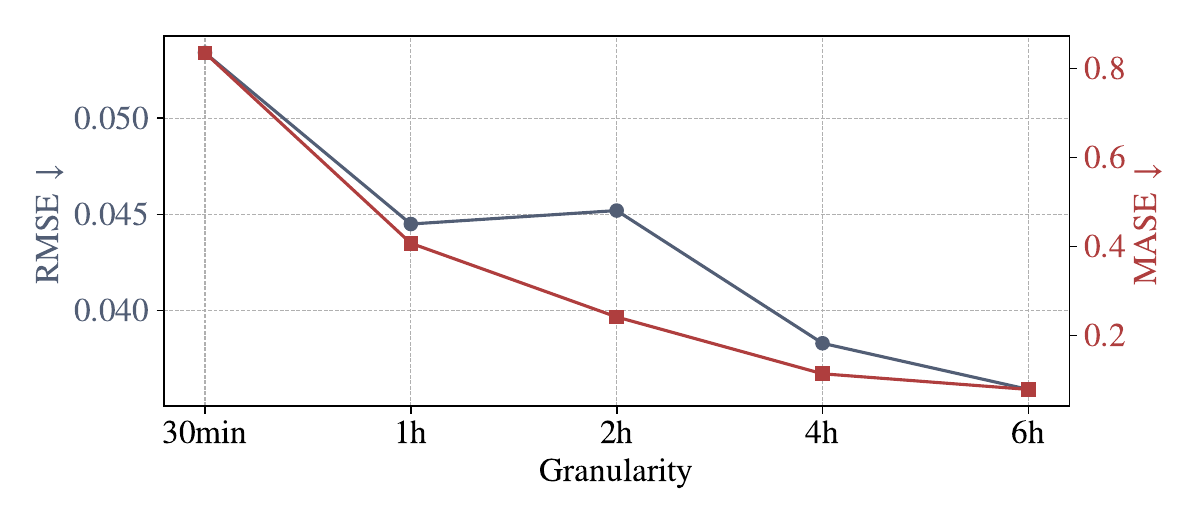}
    \caption{Evaluation of MATNet’s robustness across different time granularities, with performance reported in terms of RMSE (left axis) and MASE (right axis).}
    \label{fig:granularity}
\end{figure}

To assess the impact of data resolution, we conducted a comparative analysis across multiple input granularities ranging from 30 minutes to 6 hours (\autoref{fig:granularity}). 

To generate coarser resolutions, the data were aggregated according to the nature of each variable. 
Photovoltaic production was resampled following the same rule defined in~\autoref{eq:coarser}.
For meteorological data, extensive variables over time (e.g., precipitation) were aggregated by summation, intensive variables (e.g., temperature, humidity, pressure, dew point, wind speed, cloud cover, GHI/DNI/DHI) by arithmetic mean, and directional variables (e.g., wind direction) by circular mean, while categorical attributes (e.g., weather description) were aggregated by mode.

The results indicate that MATNet maintains stable performance across different granularities, with errors gradually decreasing at coarser resolutions due to the smoothing of short-term fluctuations. 
The 1-hour resolution emerges as the most appropriate choice, providing a favorable trade-off between forecasting accuracy and temporal detail, while aligning with operational practices in energy markets and retaining sufficient information for day-ahead forecasting.

\section{Weather description of OpenWeatherMap}
\label{app:categoricalwxfeatures}
The following lists the various levels of the categorical \emph{Weather Description} feature provided by the OpenWeatherMap API: \emph{scattered clouds}, \emph{few clouds}, \emph{broken clouds}, \emph{overcast clouds}, \emph{sky is clear}, \emph{light rain}, \emph{thunderstorm}, \emph{moderate rain}, \emph{fog}, \emph{light intensity shower rain}, \emph{mist}, \emph{haze}, \emph{heavy intensity rain}, \emph{light intensity drizzle}, \emph{shower rain}, \emph{smoke}, \emph{thunderstorm with rain}, \emph{proximity squalls}, \emph{very heavy rain}, \emph{light intensity drizzle rain}, \emph{rain and drizzle}, \emph{drizzle}.

\section{Noise Sensitivity Analysis}
\label{app:sens_noise}
\begin{figure}
    \centering
    \includegraphics[width=0.5\textwidth]{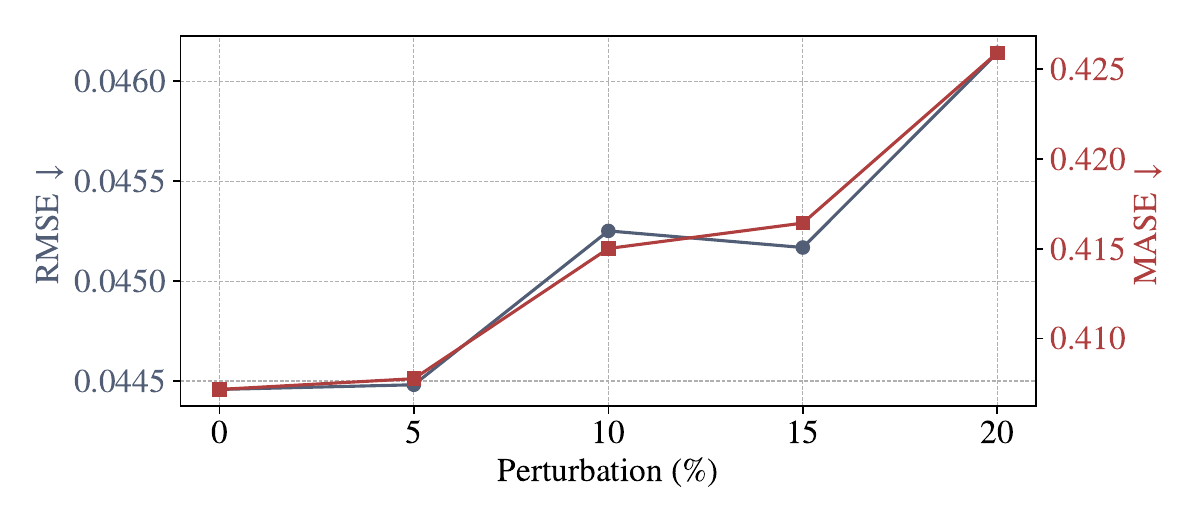}
    \caption{Noise sensitivity analysis of MATNet under increasing Gaussian perturbations (0–20 \%) applied to weather inputs, with performance reported in terms of RMSE (left axis) and MASE (right axis).}
    \label{fig:per_analysis}
\end{figure}

To evaluate the robustness of MATNet to uncertainty in weather forecasts, we conducted a sensitivity analysis by injecting Gaussian perturbations of increasing magnitude into the weather input data. Gaussian noise levels ranging from 0\% to 20\%, with 5\% increments, were applied to meteorological features to simulate potential deviations between predicted and actual weather conditions (\autoref{fig:per_analysis}). 

The results indicate that MATNet maintains stable performance across all noise levels, exhibiting only a gradual degradation as perturbations increase. Specifically, model performance remains largely consistent up to 20\% noise, with relative increases of 3.8\% in RMSE and 4.6\% in MASE compared to the noise-free condition. 
These findings confirm the robustness of MATNet against forecast uncertainty and support the realism of the 5\% Gaussian noise adopted in the main experimental setup.

\section{Baseline configurations}
\label{app:bas-config}
The statistical baselines are implemented using AutoGluon~\cite{shchur2023autogluon}.  
All models are used with default settings from AutoGluon v1.1.0.

\begin{center}
\begin{tabular}{l p{6cm}} 
\toprule
\textbf{Model} & \textbf{Default Configuration} \\
\midrule
AutoARIMA & \texttt{seasonal}=True, \texttt{stationary}=False \\
AutoCES & \texttt{model}=Z, \texttt{seasonal\_period}=None \\
NPTS & \texttt{kernel\_type}=Exponential, \texttt{use\_seasonal\_model}= True\\
Theta & \texttt{decomposition\_type}=Multiplicative, \texttt{seasonal\_period}=None \\
\bottomrule
\end{tabular}
\end{center}

The following models are specifically designed for the Ausgrid benchmark and were introduced in~\autoref{sec:relatedwork}.
All hyperparameters and model settings were selected according to the original configurations reported in the respective papers~\cite{fentis2020machine,kaur2021bayesian,kaur2022bayesian,kaur2023vae}.
\begin{center}
\begin{tabular}{l p{5cm}} 
\toprule
\textbf{Model} & \textbf{Default Configuration} \\
\midrule
LsSVR & \texttt{kernel}=Radial Basis Function\\
Bayes BiLSTM &  \texttt{optimizer}=Adam, \texttt{Loss}=KL-divergence\\
\(\alpha-\beta\) Bayes BiLSTM & \(\alpha=1.0\), \(\beta=2.0\) \\
VAE Bayes BiLSTM &\texttt{Latent dimension}=10, \texttt{Loss}=ELBO  \\
\bottomrule
\end{tabular}
\end{center}

All competitors are trained on the same splits and evaluated under identical forecasting settings as MATNet.

%\balance
\bibliographystyle{unsrtnat}
\bibliography{references.bib}

\end{document}